\documentclass[a4paper,fleqn]{cas-dc}

\usepackage[numbers]{natbib}

\usepackage{booktabs}
\usepackage{amsmath}
\usepackage{amsmath,amsfonts}
\usepackage{algorithmic}
\usepackage{algorithm}
\usepackage{array}
\usepackage{textcomp}
\usepackage{verbatim}
\usepackage{siunitx}

\def\tsc#1{\csdef{#1}{\textsc{\lowercase{#1}}\xspace}}
\tsc{WGM}
\tsc{QE}

\begin{document}
\let\WriteBookmarks\relax
\def\floatpagepagefraction{1}
\def\textpagefraction{.001}

\shorttitle{}    

\shortauthors{}  

\title [mode = title]{A Problem-Oriented Taxonomy of Evaluation Metrics for Time Series Anomaly Detection}  

\tnotemark[1] 


%

\author[1]{Kaixiang Yang}
\author[1]{Jiarong Liu}
\author[1]{Yupeng Song}
\author[2]{Shuanghua Yang}
\author[1]{Yujue Zhou\corref{cor1}}

\cortext[cor1]{Corresponding author: Yujue Zhou (email: zhouyujue@ynu.edu.cn).}

\affiliation[1]{
    organization={School of Artificial Intelligence, Yunnan University},
    addressline={},
    city={Kunming},
    postcode={650091},
    country={China}
}

\affiliation[2]{
    organization={Beijing Normal University – Hong Kong Baptist University},
    addressline={},
    city={Zhuhai},
    postcode={519087},
    country={China}
}

\fntext[funding]{
This work was supported in part by the Yunnan Fundamental Research Projects under Grant 202401AU070151, and in part by the Yunnan Provincial Science and Technology Talent and Platform Plan under Grant 202505AF350053.
}


\begin{abstract}
The rapid proliferation of IoT devices has made time series anomaly detection (TSAD) indispensable across safety-critical domains. However, existing metric taxonomies are organized by mathematical form rather than evaluation objectives, leaving practitioners without principled guidance for metric selection. This misalignment is consequential: metrics ill-suited to their context can produce misleading rankings, and in extreme cases allow random predictions to achieve scores comparable to genuine detectors.
We propose a problem-oriented framework that reinterprets over twenty metrics into six dimensions: (1) basic accuracy-driven evaluation, (2) timeliness-aware reward, (3) tolerance to labeling imprecision, (4) human-audit cost penalties, (5) robustness against random scores, and (6) parameter-free comparability. Systematic robustness experiments reveal that metrics differ by up to an order of magnitude in distinguishing genuine detectors from random guessing, and that widely adopted metrics—including NAB and Point-Adjust—show critically limited resistance to score inflation.
These findings underscore that metric suitability is inherently task-dependent, and the proposed framework provides practical guidance for selecting more context-aware and robust evaluation methodologies for TSAD in diverse IoT scenarios.
\end{abstract}


\begin{highlights}
\item Problem-oriented taxonomy organizes 20+ TSAD metrics into six functional dimensions.
\item Over 20 TSAD metrics are reformulated under a unified mathematical notation.
\item Metrics differ by up to an order of magnitude in resisting random-score inflation.
\item Practical metric selection guidelines are provided for common IoT application scenarios.
\end{highlights}



\maketitle


\section{Introduction}
The emergence of the Internet of Things (IoT) has accelerated digital transformation across numerous domains. Its defining characteristic lies in the large-scale deployment of intelligent and heterogeneous devices—such as sensors, actuators, and RFID systems—that are interconnected via the Internet to enable autonomous communication without human intervention~\cite{cook2019anomaly_lot_server}. 
In industrial contexts, the integration of IoT technologies has driven the ongoing Industry 4.0 revolution, emphasizing connectivity, automation, and intelligence. By incorporating data analytics, AI algorithms, industrial machines, and IoT sensors, Industry 4.0 aims to construct a highly efficient and intelligent industrial ecosystem.

As Industry 4.0 evolves, time series data—the foundation of intelligent monitoring and measurement—has been generated at an unprecedented rate. This surge underscores the urgent need for advanced analytical tools capable of accurately identifying anomalies within massive data streams. Beyond industrial systems, time series anomaly detection (TSAD) plays a crucial role in various application scenarios. The rapid progress of machine learning has further fueled the evolution of TSAD methods, leading to increasingly sophisticated models with complex data dependencies and adaptive learning capabilities.

However, the evaluation of anomaly detection performance has not progressed at the same rate as the development of new algorithms. The choice of evaluation metric fundamentally determines how detection behavior is quantified, interpreted, and compared under different operational requirements. In practice, different IoT and cyber-physical applications place distinct priorities on what “good performance” means: some emphasize early detection to prevent cascading failures, others focus on robustness under noisy or uncertain labels, while resource-constrained systems may prioritize metrics that reflect human-in-the-loop inspection cost. As a result, each metric implicitly reflects a particular evaluation objective that it is designed to emphasize—yet this objective is rarely made explicit in existing studies.

Despite their importance, anomaly detection metrics are seldom examined from an objective-oriented and application-driven standpoint. Existing taxonomies typically group metrics by mathematical form, aggregation level (point-based vs. event-based), or thresholding scheme. While these classifications are descriptive, they do not reveal the underlying evaluation motivations. For reliable and interpretable assessment—especially in heterogeneous IoT environments—it is crucial to understand metrics as tools designed to answer specific performance questions: What aspect of detection does the metric emphasize? Under what conditions does it behave reliably? And what types of uncertainty or bias does it introduce?

To address this gap, this paper presents an application-driven reexamination and reclassification of anomaly detection evaluation metrics. Unlike traditional taxonomies that focus primarily on computational structure, our framework organizes metrics according to their underlying evaluation objectives—the specific performance demands they are intended to capture. This objective-aware perspective enables a clearer understanding of how metrics relate to one another, how they align with diverse IoT application needs, and how they influence the interpretation of algorithmic performance. By framing anomaly detection evaluation around operational requirements, rather than purely mathematical definitions, this work offers practical and theoretically grounded guidance for selecting and designing metrics in time series anomaly detection. The main contributions of this paper are summarized as follows:
\begin{enumerate}
    
\begingroup

    \item \textbf{A problem-driven taxonomy of TSAD evaluation metrics:} \\
    We propose a new taxonomy that categorizes metrics according to the practical evaluation problems they address—such as detection timeliness, label uncertainty, threshold dependence, and inspection cost—rather than purely by their mathematical formulation. This problem-oriented perspective connects metric selection with real-world operational requirements in IoT anomaly detection.

    \item \textbf{A unified formalization of over twenty widely used metrics:} \\
    Since the original papers adopt heterogeneous notation and inconsistent definitions, we reformulate all metrics under a unified mathematical notation, enabling direct comparison of their computation rules and evaluation assumptions.
    
    \item \textbf{A systematic robustness analysis against random predictions:} \\
    We conduct controlled experiments comparing genuine detectors, random predictors, and oracle predictions, quantifying each metric’s ability to distinguish meaningful detection from random scoring.

\endgroup

    \item \textbf{Guidelines for Metric Selection and Future Metric Design:} \\
    Based on the proposed framework, we provide guidelines for selecting suitable metrics under different operational constraints, and discuss future directions for designing next-generation metrics that improve fairness, robustness, and consistency in TSAD evaluation across IoT scenarios.
\end{enumerate}

The remainder of this paper is organized as follows. Section II provides the necessary background, including the formal definition of TSAD tasks, the standard evaluation pipeline, and a review of existing metric classification approaches along with their limitations. Section III presents a unified formalization of over twenty widely used evaluation metrics, reorganized according to the proposed problem-oriented taxonomy across six functional dimensions. Section IV describes the experimental setup and reports systematic robustness experiments comparing genuine detectors, random predictors, and oracle-based attacks across both synthetic and real-world benchmarks. Section V translates the analytical findings into practical metric selection guidelines, including a capability matrix and recommended metric combinations for common application scenarios. Finally, Section VI concludes the paper and discusses directions for future metric design.

\section{Background}

This section introduces the fundamental concepts necessary for understanding the subsequent discussions, including the basic formulation of time series anomaly detection tasks, the standard evaluation pipeline, and the classification framework of evaluation metrics.

\subsection{Time Series Anomaly Detection (TSAD) and Evaluation}

\subsubsection{Task Definition and Output Form}

A time series is an ordered sequence of observations indexed by time.  In anomaly detection tasks, the focus is often placed on contiguous segments of abnormal behaviors, termed as \emph{anomalous events}.  
Accordingly, the goal of \emph{Time Series Anomaly Detection (TSAD)} is to identify these anomalous events and their corresponding anomalous points.

A large body of literature has proposed various detection approaches, ranging from statistical models to deep learning frameworks~\cite{box2013box_models,ruff2018deep_models,zong2018deep_models,xu2024calibrated_models,saravanan2023time_models,chandola2009anomaly_models,zhang2025self_lot_tsad,qin2023multiview_lot_tsad,10177932_lot_tsad,10643170_lot_tsad,chen_LearningGraphStructures_2022_lot_tsad}.  
However, this work focuses not on the detection algorithms themselves, but rather on how to \emph{evaluate} the effectiveness of their detection results.

In a typical TSAD task, a detector outputs a sequence of continuous anomaly scores:
\[
s = [s_1, s_2, \ldots, s_T],
\]
where $s_t$ denotes the confidence or probability that the time point $t$ is anomalous.  
The core objective of evaluation is to map this score sequence into an overall performance score, so as to compare the relative quality of different detectors.

\subsubsection{Evaluation Procedure and Computation Rules}

The evaluation process generally consists of three consecutive stages, as illustrated in Fig.~2:

\begin{itemize}
    \item \textbf{Score Generation:}  
    The detector outputs an anomaly score for each time point, indicating its likelihood of being anomalous.

    \item \textbf{Thresholding:}  
    A threshold $\tau$ is applied to convert the continuous score sequence into binary predictions, $\widehat{y_t} = I(s_t > \tau)$.  
    The threshold may be determined by a fixed rule, such as $\text{mean} + n \cdot \text{std}$, or by adaptive approaches, e.g., the non-parametric dynamic threshold proposed in~\cite{hundman_DetectingSpacecraftAnomalies_2018_segment-wise}.

    \item \textbf{Performance Evaluation:}  
    Once binary predictions are obtained, evaluation reduces to measuring the consistency between the detected results and the ground truth (GT) labels.
\end{itemize}

Let both the predicted results and the ground-truth labels be represented as binary time sequences, either on a point-wise, segment-wise, or sequence-wise basis.  
The standard definitions are as follows:
\begin{align*}
\mathrm{TP} &: \text{Correctly detected anomalous points or segments},\\
\mathrm{FP} &: \text{Normal parts incorrectly marked as anomalies},\\
\mathrm{FN} &: \text{Undetected anomalies},\\
\mathrm{TN} &: \text{Normal parts correctly identified as non-anomalous.}
\end{align*}

Based on these definitions, \emph{Precision}, \emph{Recall}, and \emph{F-measure} are defined as:
\[
P = \frac{TP}{TP + FP}, \quad
R = \frac{TP}{TP + FN}, \quad
F_\beta = (1+\beta^2)\frac{P \cdot R}{\beta^2 P + R},
\]
where $\beta = 1$ is commonly used to balance precision and recall.

The ground-truth anomaly labels can be represented in three equivalent forms: pointwise, segmentwise, and sequencewise. 
These respectively correspond to labeling individual timestamps, 
defining anomalous intervals, and encoding anomaly membership as a binary sequence over time.

It is worth noting that GT labels in practical scenarios are often \emph{imprecise}.  
The start and end boundaries of anomalous events can be ambiguous, and inter-annotator disagreements may lead to inconsistent GT annotations.  
Such label uncertainty can introduce substantial bias into evaluation results, thereby complicating interpretation and comparison across studies.

\subsubsection{Thresholding and Definition of Binary Evaluation Metrics}

Thresholding serves as a critical dividing step in the evaluation pipeline:

\begin{itemize}
    \item Metrics computed \emph{after} thresholding are termed \textbf{Binary Metrics}. It evaluate the combined effect of the detector and its thresholding strategy. They are based on the confusion matrix (\(TP, FP, FN, TN\)) and include classical measures such as \emph{Precision}, \emph{Recall}, and their harmonic combination, the \emph{F1-score};
    \item Metrics computed \emph{directly} from raw scores are termed \textbf{Non-binary Metrics}. In contrast, it bypasses the thresholding step and operates directly on the continuous anomaly scores.  Representative examples include AUC, VUS, and P@K. These metrics assess the detector’s global discriminative ability through ranking or integral measures over all thresholds, thereby providing a threshold-independent comparison across detectors.
\end{itemize}

In general, binary metrics emphasize operational effectiveness in deployment, while non-binary metrics focus on intrinsic discriminative capability at the model level.  
Balancing these two perspectives remains one of the core challenges in establishing a fair and meaningful evaluation system for time series anomaly detection.

\begin{figure*}[t]
    \centering
    \includegraphics[width=0.9\linewidth]{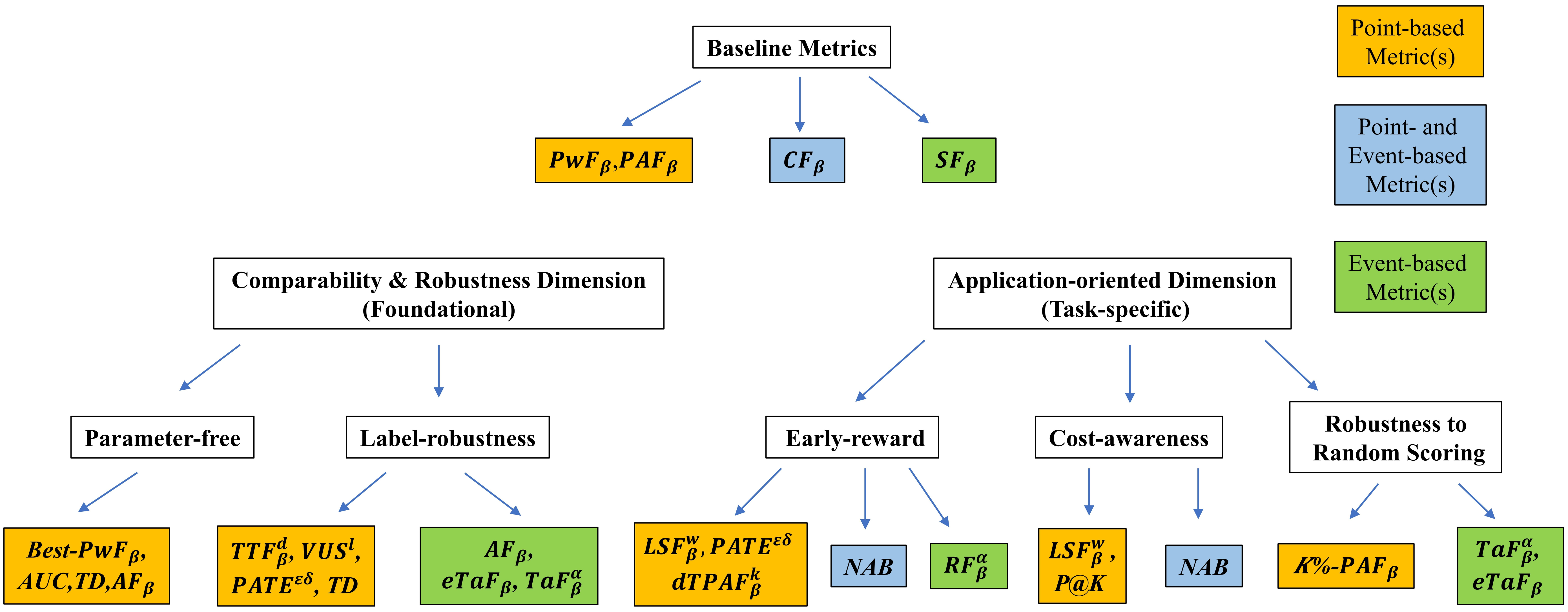}
    \caption{Problem-oriented Taxonomy of Time Series Anomaly Detection Metrics. This figure illustrates the proposed problem-oriented taxonomy of anomaly detection metrics. The framework is organized along two principal dimensions: a Foundational dimension emphasizing comparability and robustness (e.g., parameter-free and label-robust metrics), and an Application-oriented dimension reflecting task-specific evaluation goals (e.g., early-reward, cost-awareness, and robustness to random scoring). Each subcategory is associated with representative metrics, color-coded by their granularity — point-based (yellow), event-based (green), or hybrid (blue). This hierarchical structure highlights how different metrics are designed to address distinct evaluation challenges and reveals their conceptual interrelations.}
    \label{fig:main_p}
\end{figure*}

\subsection{Existing Classification Methods for Evaluation Metrics and Their Limitations}

Existing literature commonly classifies evaluation metrics based on their \emph{definition form} and \emph{computational characteristics}.  

Traditional surveys~\cite{sorbo_NavigatingMetricMaze_2024} adopt a \emph{form-based} taxonomy, where metrics are first divided into two broad groups — \textbf{binary} and \textbf{non-binary} evaluation metrics — depending on whether a thresholding step is involved.

\begin{table}[!htbp]
\centering
\caption{Table of Abbreviations and Full Names of Various Metrics}
\label{tab:abbrev}
\renewcommand{\arraystretch}{1.25}
\begin{tabular}{ll}

\toprule
\textbf{Abbreviations} & \textbf{Full Names} \\
\midrule
$PwF_{\beta}$ & Point-wise F-score~\cite{bashar_TAnoGANTimeSeries_2020_TSAD} \\
$PAF_{\beta}$ & Point Adjusted F-score~\cite{berrar_CaveatsPitfallsROC_2012_Point_Adjusted} \\
${S}F_{\beta}$ & Segment-wise F-score~\cite{hundman_DetectingSpacecraftAnomalies_2018_segment-wise} \\
$CF_\beta$ & Composite F-score~\cite{garg_EvaluationAnomalyDetection_2022_composute} \\
${dT\text{-}PAF}^k_{\boldsymbol{\beta}}$ & Delayed threshold Point Adjusted F-score~\cite{chen_JointModelIT_2021_Delay_thresholded} \\
$NAB$ & NAB Score~\cite{lavin_EvaluatingRealTimeAnomaly_2015_NAB} \\
$RF^{\alpha}_{\beta}$ & Range-based F-score~\cite{tatbul_PrecisionRecallTime_2018_range-based} \\
$TTF_\beta^\tau$ & Time tolerant F-score~\cite{scharwachter_StatisticalEvaluationAnomaly_2020_Time_tolerant} \\
$TD$ & Temporal distance~\cite{kovacs_EvaluationMetricsAnomaly_2019_Temporal_distance} \\
$AF_\beta$ & Affiliation-based F-score~\cite{huet_LocalEvaluationTime_2022_Affiliation-based} \\
$TaF_\beta^\delta$ & Time series aware F-score~\cite{hwang_TimeSeriesAwarePrecision_2019_tims_serise_aware_taf} \\
$eTaF_\beta$ & Enhanced time series aware F-score~\cite{hwang_YouKnowExisting_2022_Enhanced_time_aware_etaf} \\
$VUS^l$ & Volume Under Surface~\cite{paparrizos_VolumeSurfaceNew_2022_p@k_VUS} \\
$PATE^{\varepsilon\delta}$ & Proximity-Aware Time series Evaluation~\cite{ghorbani2024pate} \\
$\mathit{LSF}^w$ & Latency and Sparsity-aware F-score~\cite{abdulaal_PracticalApproachAsynchronous_2021_TSAD_latency_sparsity} \\
$P@K$ & Precision at K~\cite{paparrizos_TSBUADEndtoendBenchmark_2022_p@k} \\
$\mathit{K\%\text{-}PAF^k_{\beta}}$ & K\%-Point-adjusted F-score~\cite{kim_RigorousEvaluationTimeSeries_2022_k_Point_adjusted} \\
$\mathit{Best\text{-}PwF}$ & Best Threshold Binary Metrics~\cite{liu_TimeSeriesAnomaly_2023} \\
$AUC$ & Area Under Curve~\cite{feng_TimeSeriesAnomaly_2021_AUC} \\
\bottomrule
\end{tabular}

\end{table}

\begingroup

From a practical standpoint, While such classifications are useful for organizing the diverse forms of evaluation metrics, they mainly describe 
\emph{how a metric is computed} rather than 
\emph{what evaluation objective it serves}.

As a result, these taxonomies provide limited guidance for a common 
practical question faced by researchers and practitioners: 
\emph{which metric should be used under a specific application requirement?}
For example, existing structural categorizations do not explicitly reveal 
whether a metric is robust to uncertain ground-truth boundaries, 
sensitive to threshold choices, or capable of rewarding early detection.

Therefore, this work adopts a \emph{problem-oriented} classification perspective: starting from the criteria of comparability and fairness, and then categorizing metrics by their functional roles in application contexts.  This perspective shifts the focus from describing metric definitions 
to understanding their functional roles in practical TSAD evaluation.

Consequently, our framework complements existing structural taxonomies 
and provides a diagnostic tool for metric selection, 
helping practitioners match evaluation metrics to specific 
application requirements.

\endgroup

As shown in Fig.~\ref{fig:main_p}, the framework consists of two complementary dimensions(for the meanings of abbreviations, please refer to Table~\ref{tab:abbrev}). \begingroup
 To make these dimensions operational, each property is associated with a 
clear evaluative criterion that can be used to determine whether a metric 
satisfies the corresponding dimension. 
This enables systematic comparison of existing metrics and facilitates 
the design of new evaluation metrics tailored to specific application needs:
\endgroup

\textbf{Comparability \& Robustness Dimension (Foundational):}

\begingroup

\begin{itemize}
    \item \textbf{Threshold-dependence / Parameter-free:} Indicates whether the metric requires user-defined or data-dependent 
    hyperparameters (e.g., thresholds or tolerance windows) to convert 
    continuous anomaly scores into binary decisions.
    
    Operationally, a metric is considered \emph{threshold-dependent} if its 
    final score cannot be computed directly from continuous anomaly scores 
    and ground-truth labels without specifying additional parameters. 
    Conversely, \emph{parameter-free} metrics can be evaluated directly from 
    the ranking or distribution of anomaly scores without introducing such 
    hyperparameters.
    
    \item \textbf{Label-robustness:} Measures the stability of a metric under uncertainty or inconsistency 
    in ground-truth anomaly boundaries.
    
    Operationally, a metric is considered \emph{label-robust} if small 
    perturbations to the annotated anomaly boundaries (e.g., slight shifts 
    or expansions of anomaly intervals) do not cause substantial changes 
    in the resulting evaluation score. 
    This property is particularly important in industrial anomaly detection 
    settings where event boundaries are often imprecise or subject to 
    human labeling variability.
\end{itemize}

\textbf{Application-oriented Dimension (Task-specific):}
\begin{itemize}
    \item \textbf{Timeliness / Early-reward:} A metric is considered \emph{timeliness-aware} if it assigns higher rewards to detections that occur earlier within an anomaly interval.
    
    Operationally, given an anomaly segment $[t_s, t_e]$, a metric exhibits timeliness if the reward assigned to a detection at time $t$ decreases as $t$ moves away from the start of the anomaly window. 
    This property ensures that earlier detections receive greater credit than later ones.
    
    \item \textbf{Cost-awareness / Sparse-alarm penalty:} A metric is considered \emph{cost-aware} if its scoring function reflects 
    the asymmetric operational impact of different prediction outcomes, 
    such as false alarms and missed detections.
    
    Operationally, this property is present when the metric introduces 
    explicit weighting or penalty mechanisms that discourage excessive 
    false alarms or reward more reliable detections, thereby aligning the 
    evaluation with practical operational costs in real-world monitoring systems.
    
    \item \textbf{Score-stability / Robustness to random scoring:} A metric is considered \emph{score-stable} if its design discourages 
    artificially inflated evaluation scores produced by sparse or random 
    prediction strategies.
    
    In practice, robustness to random or sparse scoring typically emerges 
    when a metric explicitly constrains what qualifies as a valid anomaly 
    detection. Overall, these metrics share a common design philosophy: a detection is considered valid only if it demonstrates sufficient temporal 
    consistency with the ground-truth anomaly segment.
    
    Empirical analysis with random prediction baselines can then be used as a 
    diagnostic tool to examine whether the metric exhibits this desirable 
    property in practice.
\end{itemize}

Unlike traditional taxonomies that assign each metric to a single structural category, 
the proposed framework allows a metric to be associated with multiple dimensions 
simultaneously. 
For instance, a metric may exhibit strong timeliness properties while still being 
sensitive to threshold selection.

This multi-dimensional perspective enables the framework to function as a 
diagnostic tool for evaluation design. 
In practice, engineers can first examine the foundational properties of a metric 
(e.g., threshold dependence and label robustness), and then select metrics that 
align with specific application requirements such as early detection or 
false-alarm cost control.

Together, these analyses illustrate how the proposed taxonomy can support 
more transparent and task-aware evaluation practices in TSAD research.

\endgroup

\section{Detailed Metrics}

\subsection{Baseline Metrics}
\subsubsection{\textbf{Formal Definitions}}
~\
\paragraph{Point-wise F-score ($PwF_{\beta}$)}
~\\
A basic evaluation paradigm treats every timestamp in a time series as an independent classification instance, i.e., deciding whether the point at time $t$ is anomalous. The advantages of point-wise metrics are simplicity, interpretability, and direct computability, which explains their widespread usage in early TSAD research~\cite{bashar_TAnoGANTimeSeries_2020_TSAD,han_LearningSparseLatent_2022_TSAD,li_MultivariateTimeSeries_2021_TSAD,abdulaal_PracticalApproachAsynchronous_2021_TSAD_latency_sparsity}.

Formally, let the time series length be $T$, the ground-truth label at time $t$ be $y_t \in \{0,1\}$ and the predicted binary label be $\hat{y}_t \in \{0,1\}$. Define
\[
\mathrm{TP} = \sum_{t=1}^T y_t \hat{y}_t, \quad
\mathrm{FP} = \sum_{t=1}^T (1 - y_t)\hat{y}_t, \quad
\mathrm{FN} = \sum_{t=1}^T y_t (1 - \hat{y}_t).
\]
Precision, Recall and $F_\beta$ are then computed in the usual way:
\[
P = \frac{\mathrm{TP}}{\mathrm{TP} + \mathrm{FP}}, \quad
R = \frac{\mathrm{TP}}{\mathrm{TP} + \mathrm{FN}}, \quad
F_\beta = (1+\beta^2)\frac{P\cdot R}{\beta^2 P + R}.
\]

\paragraph{Point Adjusted F-score (PA$F_{\beta}$)}
~\\
To mitigate the excessive strictness of point-wise metrics, the \emph{point-adjusted} approach was proposed~\cite{berrar_CaveatsPitfallsROC_2012_Point_Adjusted}. The basic idea is simple and application-driven: if a model predicts at least one positive point within a ground-truth anomalous interval, that entire interval is considered ``detected''. This reflects the practical logic that, in some scenarios, a single alert is sufficient to trigger a human investigation.

Concretely, let the set of ground-truth anomalous segments be
\[
\mathcal{G} = \{[s_i, e_i]\}_{i=1}^{N_g},
\]
and let the original binary predictions be $\{\hat{y}_t\}_{t=1}^T$. The point-adjusted prediction $\tilde{y}_t$ is obtained by adjusting predictions within each ground-truth segment as follows:

\[
\begin{aligned}
\tilde{y}_t = \begin{cases}
1, & \text{if } \exists\, i \text{ s.t. } t \in [s_i, e_i] \text{ and } \\
   & \exists\, u \in [s_i, e_i] \text{ with } \hat{y}_u = 1, \\
\hat{y}_t, & \text{otherwise}.
\end{cases}
\end{aligned}
\]

\paragraph{Segment-wise F-score ($SF_{\beta}$)}
~\\
Segment-wise evaluation treats each continuous anomalous interval as an individual event, and counts whether each event is detected or missed~\cite{hundman_DetectingSpacecraftAnomalies_2018_segment-wise,flaborea2023we_SF,geiger2020tadgan_SF,meng2019spacecraft_SF,nalepa2022evaluating_SF}. Let the ground-truth segments be
\[
\mathcal{G}=\{G_i=[s_i,e_i]\}_{i=1}^{N_g},
\]
and the predicted segments be
\[
\mathcal{P}=\{P_j=[s_j',e_j']\}_{j=1}^{N_p}.
\]
Define the overlap predicate between two intervals as
\[
\mathrm{overlap}([s_1,e_1],[s_2,e_2])\colon= \neg(e_1 < s_2 \vee e_2 < s_1),
\]
equivalently $s_1 \le e_2$ and $s_2 \le e_1$.

The confusion counts for segment-level evaluation are then:
\begin{align*}
\mathrm{TP}_{\mathrm{seg}} &= \bigl|\{G_i\in\mathcal{G}\mid \exists\,P_j\in\mathcal{P},\ \mathrm{overlap}(G_i,P_j)\}\bigr|, \\
\mathrm{FN}_{\mathrm{seg}} &= \bigl|\{G_i\in\mathcal{G}\mid \forall\,P_j\in\mathcal{P},\ \neg\mathrm{overlap}(G_i,P_j)\}\bigr|, \\
\mathrm{FP}_{\mathrm{seg}} &= \bigl|\{P_j\in\mathcal{P}\mid \forall\,G_i\in\mathcal{G},\ \neg\mathrm{overlap}(P_j,G_i)\}\bigr|.
\end{align*}
Precision and Recall at the event (segment) level are
\[
P_{\mathrm{seg}} = \frac{\mathrm{TP}_{\mathrm{seg}}}{\mathrm{TP}_{\mathrm{seg}} + \mathrm{FP}_{\mathrm{seg}}}, \qquad
R_{\mathrm{seg}} = \frac{\mathrm{TP}_{\mathrm{seg}}}{\mathrm{TP}_{\mathrm{seg}} + \mathrm{FN}_{\mathrm{seg}}},
\]
and the Segment-wise $F_\beta$ is computed as
\[
SF_\beta = (1+\beta^2)\frac{P_{\mathrm{seg}}\cdot R_{\mathrm{seg}}}{\beta^2 P_{\mathrm{seg}} + R_{\mathrm{seg}}}.
\]

\paragraph{Composite F-score (CF$_\beta$)}
~\\
Event-level metrics and point-wise metrics are complementary: the former emphasize event discovery (recall at event level) while the latter emphasize localization precision. To balance these two perspectives, the \emph{composite F-score} ($\mathrm{CF}_\beta$) integrates an event-level recall with a point-wise precision via a harmonic combination~\cite{garg_EvaluationAnomalyDetection_2022_composute}.

Specifically, let event-level recall be
\[
R_{\mathrm{seg}} = \frac{\mathrm{TP}_{\mathrm{seg}}}{\mathrm{TP}_{\mathrm{seg}} + \mathrm{FN}_{\mathrm{seg}}},
\]
and point-wise precision be
\[
P_{\mathrm{point}} = \frac{\mathrm{TP}_{\mathrm{point}}}{\mathrm{TP}_{\mathrm{point}} + \mathrm{FP}_{\mathrm{point}}},
\]
where $\mathrm{TP}_{\mathrm{point}}, \mathrm{FP}_{\mathrm{point}}$ are the standard point-wise counts (as defined in the point-wise subsection). The composite F-score is then
\[
CF_\beta = (1+\beta^2)\frac{P_{\mathrm{point}} \cdot R_{\mathrm{seg}}}{\beta^2 P_{\mathrm{point}} + R_{\mathrm{seg}}}.
\]

\begingroup

Baseline metrics differ primarily in the unit of correctness they adopt, which directly determines how confusion counts are constructed and how detection behavior is rewarded or penalized. The granularity of the evaluation reflects different objectives of anomaly detection, namely the trade-off between precise temporal localization and event-level discovery.

\endgroup

\subsection{Timeliness and Early Detection Metrics}

In many time-critical domains—such as industrial fault prevention, medical monitoring, or financial risk control—the \textit{timeliness} of anomaly detection becomes a key evaluation dimension. 
Traditional point- or segment-level F-scores only measure whether anomalies are correctly identified but remain insensitive to \textit{when} they are detected. 
Consequently, they fail to capture the operational value of early alarms, where even a short advance warning can substantially mitigate downstream losses. 
To address this limitation, a series of \textit{timeliness-aware} metrics have been proposed to explicitly reward earlier detections while penalizing delayed or redundant alerts.

\subsubsection{\textbf{Formal Definitions}}
~\
\paragraph{Delayed threshold Point Adjusted F-score (\( dT\text{-}PAF^k_{\boldsymbol{\beta}} \))}

This metric assumes that if a model raises an alarm within the first \( k \) time steps of a true anomaly segment, the anomaly is considered ``timely captured.'' Even if later predictions deviate, the segment is still treated as successfully detected. Conversely, if the detection delay exceeds \( k \), the entire segment is regarded as a miss, thereby penalizing late detections~\cite{chen_JointModelIT_2021_Delay_thresholded}. This better aligns with real-world scenarios where early detection can prevent major losses. However, it may overestimate precision for long anomaly segments.

Such a design highlights the model’s responsiveness — in applications where ``early discovery'' is of paramount importance, it effectively distinguishes models with stronger temporal sensitivity. Meanwhile, by restricting detections to the first \( k \) steps, it partially mitigates the impact of random hits in long anomaly segments. Nonetheless, the choice of \( k \) remains subjective; the definition of ``timely'' may vary across domains, limiting cross-task comparability.

\textbf{Computation.}
Let the true anomaly segment be \( [s, e] \).
If there exists any predicted point \( \hat{y}_t = 1 \) within the window \( [s, s+k-1] \) (where \( s+k-1 \leq e \)), the entire segment is marked as detected:
\[
y_t = 1, \quad \forall t \in [s, e].
\]
Otherwise, all predicted points within this segment are removed and counted as false negatives (FN).
After this adjustment, the point-wise counts \( \mathrm{TP}, \mathrm{FP}, \mathrm{FN} \) are recalculated, and the score is defined as:
\[
dT\text{-}PAF^k_{\beta} =
\frac{(1+\beta^2)\cdot \mathrm{TP}}{(1+\beta^2)\cdot \mathrm{TP} + \beta^2\cdot \mathrm{FN} + \mathrm{FP}},
\]
where \( \beta \) controls the trade-off between precision and recall.

\paragraph{NAB Score}
~\\
To address the same timeliness requirement, Numenta proposed the \textit{Numenta Anomaly Benchmark (NAB) Score}, a widely recognized framework that integrates \textit{detection accuracy}, \textit{timeliness}, and \textit{false alarm sparsity}~\cite{lavin_EvaluatingRealTimeAnomaly_2015_NAB, paparrizos2022tsb_nab,schmidl2022anomaly_nab,lavin2015evaluating_nab}.

The NAB design revolves around \textit{temporal windows} and \textit{dynamic weighting}. For each true anomaly segment, a detection window of equal length is defined. If a prediction falls within this window, it receives a reward based on its position — earlier detections are rewarded more, while delayed detections are penalized.

Specifically, the reward function is modeled by a scaled sigmoid:
\[
S(r) =
\begin{cases}
-1, & r > 3.0, \\
2 \cdot \mathrm{sigmoid}(-5r) - 1, & \text{otherwise},
\end{cases}
\]
where:
\begin{itemize}
    \item For detections \textbf{inside} the window:
    \( r = -\frac{R_i - t + 1}{W} \),
    with \( R_i \) the right boundary of the window, \( t \) the detection time, and \( W \) the window width.
    \item For \textbf{false alarms} outside the window:
    \( r = \frac{|t - R_{\text{prev}}|}{W_{\text{prev}} - 1} \),
    where \( R_{\text{prev}} \) and \( W_{\text{prev}} \) denote the boundary and width of the nearest preceding window.
\end{itemize}

Each window contributes at most one positive reward (the maximum within-window score). False alarms accumulate negative scores:
\[
\mathrm{RawScore} =
\sum_i \max_{t \in W_i}\left(w_{tp} \cdot S(r_t)\right)
- \sum_j w_{fp} \cdot S(r_j),
\]
where \( w_{tp} \) and \( w_{fp} \) are weighting factors for true and false detections.

The final NAB score is normalized as:
\[
\mathrm{NAB} = 100 \cdot
\frac{\mathrm{RawScore} - \mathrm{NullScore}}{\mathrm{PerfectScore} - \mathrm{NullScore}},
\]
where \textit{NullScore} represents the baseline (no detections) and \textit{PerfectScore} corresponds to ideal performance.

\paragraph{Range-based F-score (\( RF^{\alpha}_{\beta} \))}
~\\
In complex real-world scenarios, binary hit/miss outcomes — even at segment-level — often fail to fully capture detection quality. Different tasks prioritize different goals: some emphasize \textit{early detection}, others \textit{comprehensive coverage}, and others wish to avoid \textit{redundant detections}. To unify these perspectives, researchers proposed the \textbf{Range-based F-score} (\( RF^{\alpha,\beta} \)), a flexible metric framework that integrates various reward and penalty mechanisms through customizable functional modules~\cite{tatbul_PrecisionRecallTime_2018_range-based,gensler2014novel_RF,jacob2020exathlon_RF,meng2019spacecraft_RF}.

The key idea is to decompose traditional \textit{precision} and \textit{recall} into modular components:
\begin{enumerate}
    \item \textbf{Existence function} — determines whether a true or predicted interval is hit at least once;
    \item \textbf{Coverage function} — quantifies the degree of overlap;
    \item \textbf{Cardinality function} — penalizes multiple detections of the same true interval;
    \item \textbf{Positional function} — weights detections based on temporal preferences (e.g., early vs. late).
\end{enumerate}

By tuning these modules and their parameters, the metric can adapt to diverse application objectives. For instance, in security monitoring, the positional function can emphasize ``earlier is better,'' while in industrial quality control, the coverage term can dominate to ensure \textit{complete detection}. This modularity grants \( RF^{\alpha,\beta} \) high interpretability and task adaptability.

However, the cardinality penalty \( g(n) \) (often \( g(n)=1/n \)) can introduce excessive subjectivity. Its original intent is to prevent \textit{artificial inflation} — e.g., splitting one true anomaly into many small predictions to boost recall — but it may overly penalize slightly fragmented yet legitimate detections.

For example, when a model produces multiple short but adjacent detections for the same anomaly, the overall coverage may still be sufficient, yet the score is forcibly compressed by \( 1/n \). This bias favors \textit{compact} detections, which is suitable for single-alarm settings but distorts evaluation for continuous monitoring or multi-level alert systems. Hence, careful tuning of the penalty and positional functions is recommended.

\textbf{Computation.} Let \( I_r \) and \( I_p \) denote the sets of true and predicted anomaly intervals.
For any \( i \in I_r, j \in I_p \), let \( \mathrm{overlap}(i, j) \) denote their intersection length and \( |i| \) the length of interval \( i \).

\begin{itemize}
    \item \textbf{Existence}
    \[
    E_i^r = \mathbb{I}\{\exists j \in I_p : \mathrm{overlap}(i, j) > 0\},
    \]
    \[
    E_j^p = \mathbb{I}\{\exists i \in I_r : \mathrm{overlap}(i, j) > 0\}.
    \]

    \item \textbf{Coverage}
    \[
    \mathrm{cov}(i) = \frac{\sum_j \mathrm{overlap}(i, j)}{|i|}, \quad
    C_i^r = \mathrm{Transform}(\mathrm{cov}(i)),
    \]
    where \textit{Transform} can be identity, truncation, or another monotonic function.

    \item \textbf{Positional Weight}
    \[
    \omega_{ij}^r \in [0,1],
    \]
    representing the relative importance of overlap location within \( i \), controlled by shape parameters such as \textit{flat}, \textit{early}, or \textit{late}. Normalized forms satisfy \( \sum_j \omega_{ij}^r = 1 \).

    \item \textbf{Cardinality Penalty}
    \[
    n_i = |\{ j \in I_p : \mathrm{overlap}(i, j) > 0 \}|, 
    \]
    \[
    g(n_i)\ \text{is a non-increasing function (e.g., } g(n)=1 \text{ or } 1/n \text{)}.
    \]
\end{itemize}

Then, the weighted contribution of each true interval is:
\[
S_i^r = \alpha \cdot E_i^r + (1-\alpha) \cdot g(n_i) \cdot \sum_j \omega_{ij}^r.
\]

The recall and precision are computed as:
\[
R^\alpha = \frac{1}{|I_r|} \sum_{i \in I_r} S_i^r, \quad
P^\alpha = \frac{1}{|I_p|} \sum_{j \in I_p} S_j^p.
\]

Finally, the overall range-based F-score is given by:
\[
RF_\beta^\alpha =
\frac{(1+\beta^2) P^\alpha R^\alpha}{\beta^2 P^\alpha + R^\alpha}.
\]

\paragraph{Latency and Sparsity-aware F-score (\(\mathit{LSF}^w\))}
~\\
The \(\mathit{LSF}^w\) metric implements an early-detection reward and late-detection penalty through a \emph{detection state propagation} mechanism.  
Specifically, when the model successfully detects at least one true anomaly within a window \(W_i\)—that is,
\[
\exists\, t\in W_i:\;\hat{y}_t=1 \land y_t=1,
\]
the window enters the \emph{detection-activated state}.  
In this case, LSF propagates the detection state from the first true anomalous point  
\[
s_i=\min\{t\in W_i : y_t=1\}
\]
to the end of the window, forcing all predictions in this interval to be positive:
\[
\hat{y}_t^\ast = 1,\quad 
\forall\, t \in [s_i,\, \max(W_i)],\quad
\text{if } \exists\,\tau\in W_i:\hat{y}_\tau=1\land y_\tau=1.
\]

This design embodies two layers of evaluation logic:

1. \textbf{Reward for early detection.}  
   If the model raises an alarm near the beginning of an anomalous segment (i.e., close to \(s_i\)), all subsequent points inherit correct predictions without requiring additional detections.  
   This provides a positive incentive for “fast response”—the earlier the detection, the larger the implicit trust and the higher the recall.

2. \textbf{Penalty for delayed detection.}  
   Conversely, if the model detects the anomaly only near the end of the window (e.g., close to \(\max(W_i)\)), all earlier missed points  
   \[
   t\in [s_i,\, \tau),\qquad \tau=\text{first detected point},
   \]
   remain counted as false negatives.  
   Thus, delayed detections cannot obtain full batch-level trust propagation, yielding significantly lower scores than timely detection.

In addition, LSF introduces cross-window persistence of the detection state: once a window becomes detection-activated, the state is propagated into the next window as long as ground-truth anomalies remain continuous (i.e., no \(y_t=0\) is encountered).  
This ensures coherent evaluation of long anomalous segments and avoids artificial penalties caused by window boundaries.  
However, the detection state is immediately reset when a normal point appears, forcing the model to re-demonstrate its detection capability at the beginning of each new anomalous segment, thereby maintaining evaluation strictness.

\textbf{Computation.} Given a window size \(\mathit{w} \in \mathbb{Z}^+\), the calculation proceeds as follows:

The first step is window partition. The sequence is divided into batches:
  
  \[
  W_i = [iw + 1, \min((i+1)w, T)],
  \]
  \[i = 0, 1, \ldots, B-1, \quad B = \lceil T / w \rceil.\]
  
And the next step is delay adjustment. For each window \(W_i\), define the first anomaly as:
  \[
  s_i =
  \begin{cases}
      \min\{ t \in W_i : y_t = 1 \}, otherwise,\\
      \infty, \text{ if no anomaly exists.}
      
  \end{cases}
  \]

Define a detection-state flag \(\mathrm{DS}_i\in\{0,1\}\) updated by:
\[
\mathrm{DS}_i =
\begin{cases}
1, & \begin{aligned}
    &\text{if }\exists\, t\in W_i:\hat{y}_t=1\land y_t=1 \\
    & \;\land\; (\mathrm{DS}_{i-1}=1 \;\lor\; t=s_i),\\[4pt]
    \end{aligned} \\
0, & \text{if }\forall\, t\in W_i: y_t=0.
\end{cases}
\]

The adjusted prediction sequence is:

    \[
    \hat{y}_t^\ast=
    \begin{cases}
    1, & t\in [s_i,\, \max(W_i)] \;\land\; \mathrm{DS}_i=1,\\[4pt]
    \hat{y}_t, & \text{otherwise}.
    \end{cases}
    \]
    
The batch-adjusted ground-truth labels are:
  
  \[
  y_t^* =
  \begin{cases}
    1, & t \in [s_i, \max(W_i)], \\
    0, & \text{otherwise.}
  \end{cases}
  \]

  For each window, batch-level Confusion Matrix :
  \[
  \mathrm{TP}_i = \mathbb{I}\left[\exists t \in W_i : y_t^* = 1 \land \hat{y}_t^* = 1\right], \]
  \[
  \mathrm{FP}_i = \mathbb{I}\left[\exists t \in W_i : y_t^* = 0 \land \hat{y}_t^* = 1\right], \]
  \[
  \mathrm{FN}_i = \mathbb{I}\left[\exists t \in W_i : y_t^* = 1 \land \hat{y}_t^* = 0\right],
  \]
  where \(\mathbb{I}[\cdot]\) denotes the indicator function.
  
Global aggregation and metric computation:
  \[
  \mathrm{TP} = \sum_{i=0}^{B-1} \mathrm{TP}_i, \quad
  \mathrm{FP} = \sum_{i=0}^{B-1} \mathrm{FP}_i, \quad
  \mathrm{FN} = \sum_{i=0}^{B-1} \mathrm{FN}_i.
  \]

  \[
  \mathit{LSF}^w = \frac{2\mathrm{TP}}{2\mathrm{TP} + \mathrm{FP} + \mathrm{FN}}.
  \]

\begingroup

\paragraph{Proximity-Aware Time series Evaluation ($PATE^{\varepsilon\delta}$)}
~\\
The $PATE^{\varepsilon\delta}$ metric addresses four major limitations of traditional metrics—temporal blindness, binary decisioning, delay insensitivity, and weak practical relevance—by introducing proximity-weighted regions and adaptive buffer zones~\cite{ghorbani2024pate}.

Specifically, $PATE^{\varepsilon\delta}$ defines three semantic regions around each true anomaly segment—pre-buffer, true, and post-buffer—characterized by parameters \( \varepsilon \) (early-detection buffer) and \( \delta \) (delayed-detection buffer). Predictions are weighted according to their temporal proximity to true anomalies, enabling graded evaluation of “near misses” and “early warnings.”

\textbf{Computation.} For each true anomaly segment \( A_i = [a_i^s, a_i^e] \):
\[
\mathcal{R}_i^{pre} = [\max(0, a_i^s - \varepsilon, \mathcal{R}_{i-1}^{post}+1), a_i^s - 1], \quad
\mathcal{R}_i^{true} = [a_i^s, a_i^e], 
\]
\[
\mathcal{R}_i^{post} = [a_i^e + 1, \min(a_i^e + \delta, a_{i+1}^s - 1, T)].
\]
Predictions are then classified as belonging to one of the regions: \emph{true detection}, \emph{pre-buffer}, \emph{post-buffer}, \emph{partial miss}, or \emph{outside}.

\textbf{Weighted Scoring.}  
Weights are assigned to predictions according to their distance from the true segment:
\[
w_{\mathrm{pre}}(t, A_i) = 1 - \frac{\sum_{k=a_i^s}^{a_i^e} |k - t|}{\sum_{k=a_i^s}^{a_i^e} |k - \max(0, a_i^s - \varepsilon)|},
\]
\[
w_{\mathrm{post}}(t, A_i) = 1 - \frac{\sum_{k=a_i^s}^{a_i^e} |t - k|}{\sum_{k=a_i^s}^{a_i^e} |\min(a_i^e + \delta, T) - k|}.
\]

Finally, weighted precision and recall are computed as:
\[
\mathrm{Precision} = \frac{W_{TP}}{W_{TP} + W_{FP}}, \quad
\mathrm{Recall} = \frac{W_{TP}}{W_{TP} + W_{FN}},
\]
where \( W_{TP}, W_{FP}, W_{FN} \) denote weighted true positives, false positives, and false negatives respectively.

$PATE^{\varepsilon\delta}$ supports two evaluation modes:
\[
\mathrm{PATE^{\varepsilon\delta}} = \frac{1}{|\mathcal{B}_\varepsilon||\mathcal{B}_\delta|} \sum_{(\varepsilon', \delta') \in \mathcal{B}_\varepsilon \times \mathcal{B}_\delta} \mathrm{AUC\!-\!PR}(\theta_k),
\]
\[
\mathrm{PATE^{\varepsilon\delta}\!-\!F1} = \frac{1}{|\mathcal{B}_\varepsilon||\mathcal{B}_\delta|} \sum_{(\varepsilon', \delta')} \frac{2 \cdot \mathrm{Precision} \cdot \mathrm{Recall}}{\mathrm{Precision} + \mathrm{Recall}}.
\]

\endgroup

\begingroup

\subsubsection{\textbf{Conceptual Characteristics}}
~\\
Although all timeliness-aware metrics aim to reward early detection, they implement this objective through fundamentally different mechanisms, including threshold-based criteria ($dT\text{-}PAF^k_{\beta}$), continuous reward shaping (NAB), structural preference integration (\( RF^{\alpha}_{\beta} \), \(\mathit{LSF}^w\)), and proximity-aware tolerance modeling (\(PATE^{\varepsilon\delta}\)). 
These mechanisms reflect distinct modeling philosophies of temporal sensitivity, each introducing specific inductive biases in how early or delayed detections are valued. 
As a result, model ranking can vary significantly depending on whether a metric emphasizes strict timeliness, smooth temporal reward, or tolerance to temporal misalignment. 
Moreover, such designs inevitably rely on task-dependent parameterization (e.g., thresholds, window sizes, or buffer regions), which improves operational flexibility but reduces cross-task comparability. 
From the perspective of our taxonomy, these metrics extend the timeliness and cost-awareness dimensions, at the expense of increased configuration complexity.

\endgroup

\subsection{Label-robust Metrics (Mitigating Annotation Uncertainty)}

In real-world applications, anomaly annotations often contain uncertainties. The start or end boundaries of anomalies may slightly deviate from their actual occurrence time due to human judgment or noise contamination. Consequently, traditional point-wise metrics can suffer severe score degradation caused by minor temporal misalignment, leading to an unfair evaluation of detection ability. To address this, several label-robust metrics have been proposed to tolerate annotation uncertainty and boundary fuzziness.

\subsubsection{\textbf{Formal Definitions}}
~\
\paragraph{Time tolerant F-score ($TTF_\beta^\tau$)}
~\\
The core idea of the time-tolerant F-score is to introduce a temporal tolerance radius \( d \), such that a prediction is considered correct as long as it occurs within \( d \) time steps of a true anomaly~\cite{scharwachter_StatisticalEvaluationAnomaly_2020_Time_tolerant}.

\textbf{Computation.}  
For each predicted anomaly point \( \widehat{y_t} = 1 \), if there exists a true anomaly point \( a \) satisfying \( |t - a| \le d \), it is counted as a true positive. The time-tolerant precision and recall are defined as:

\[R_d = \frac{1}{|E|} \sum_{t: y_t=1} \mathbb{I}\left(\sum_{j=t-d}^{t+d} \widehat{y_j}=1\right), \]

\[P_d = \frac{1}{|A|} \sum_{t: \widehat{y_t}=1} \mathbb{I}\left(\sum_{j=t-d}^{t+d} y_j=1\right)\]

where \( d \) is the tolerance radius, \( E \) the set of true anomalies, and \( A \) the set of predicted anomalies.  
Then the time-tolerant F-score is given by:

\[
TTF^d_{\beta} = \frac{(1+\beta^2) P_d R_d}{\beta^2 P_d + R_d}.
\]

\paragraph{Temporal distance (TD)}
~\\
To further quantify the temporal offset between predicted and true anomalies, the \emph{Temporal distance (TD)} metric adopts a bidirectional nearest-neighbor distance strategy~\cite{kovacs_EvaluationMetricsAnomaly_2019_Temporal_distance}. For each true anomaly, the minimum temporal distance to any predicted point is computed, and vice versa. The final TD score sums both distances, measuring the localization accuracy along the time axis—smaller TD values indicate better temporal precision.

\textbf{Computation.}  
Let \( A = \{ s : y_s = 1 \} \) and \( B = \{ t : \widehat{y_t} = 1 \} \) denote the sets of true and predicted anomaly points, and \( L \) the sequence length. Then:

\begin{align}
\label{TD}
\mathrm{TD} = \sum_{s \in A} d(s, B) + \sum_{t \in B} d(t, A),
\end{align}

where
\[
d(s, B) =
\begin{cases}
\min_{t \in B} |s - t|, & \text{if } B \neq \emptyset,\\
L, & \text{if } B = \emptyset,
\end{cases}
\]
\[
d(t, A) =
\begin{cases}
\min_{s \in A} |t - s|, & \text{if } A \neq \emptyset,\\
L, & \text{if } A = \emptyset.
\end{cases}
\]
This design ensures consistent penalization and robustness even when either the prediction or ground truth is empty.

\paragraph{Affiliation-based F-score ($AF_\beta$)}
~\\
The affiliation-based F-score alleviates label uncertainty by quantifying the distance-based closeness between predicted and true anomalies~\cite{huet_LocalEvaluationTime_2022_Affiliation-based}. Instead of requiring strict overlap, it transforms temporal proximity into a probabilistic affinity score, offering interpretability and robustness against boundary fuzziness.

\textbf{Computation.} The computation involves three steps.  
First, for each true anomaly segment \( \mathcal{J} = \{ J_1, \ldots, J_m \} \), define its corresponding membership region:
\[
E_i = \left( \frac{t_{\text{stop}}^{i-1} + t_{\text{start}}^i}{2}, \frac{t_{\text{stop}}^i + t_{\text{start}}^{i+1}}{2} \right).
\]
Next, project each predicted segment \( I \in \mathcal{I} = \{ I_1, \ldots, I_n \} \) onto the associated membership region:
\[
\mathcal{P}_i = \{ I \cap E_i : I \in \mathcal{I}, I \cap E_i \neq \emptyset \}.
\]
Then, compute the Affiliation-based precision and recall within each region via integration:
\[
P_{AF} = \frac{\sum_{i=1}^m \sum_{I \in \mathcal{P}_i} \int_{I} p_{\mathrm{precision}}(x, J_i, E_i) \, dx}{\sum_{I \in \mathcal{I}} |I|}, 
\]
\[
R_{AF} = \frac{\sum_{i=1}^m \int_{J_i} p_{\mathrm{recall}}(x, \mathcal{P}_i, E_i) \, dx}{\sum_{i=1}^m |J_i|}.
\]
Finally, aggregate them using:
\[
AF_\beta = (1+\beta^2) \frac{P_{AF} R_{AF}}{\beta^2 P_{AF} + R_{AF}}.
\]

\paragraph{Time series aware F-score ($TaF_\beta^\delta$)}
~\\
The time series aware F-score introduces both a coverage threshold and a fuzzy tolerance zone to balance temporal robustness and realistic event coverage. It decomposes the detection performance into two dimensions: \emph{detection accuracy} (event-level recognition) and \emph{positional accuracy} (temporal alignment)~\cite{hwang_TimeSeriesAwarePrecision_2019_tims_serise_aware_taf}.

\textbf{Computation.} Let the recall and precision be defined as:

\begin{align}
\label{TaF_2}
R = \alpha R_{\mathrm{detection}} + (1-\alpha) R_{\mathrm{position}}, \\
P = \alpha P_{\mathrm{detection}} + (1-\alpha) P_{\mathrm{position}},
\end{align}
where \( \alpha \in [0,1] \) balances event-level and positional emphasis.

\textbf{Detection-level components:}

\begin{align}
\label{TaF_1}
R_{\mathrm{detection}} &= \frac{1}{|A|} \sum_{i=1}^{|A|} \mathbb{I}\!\left(\frac{S_i}{L_i} \ge \theta\right), \\
P_{\mathrm{detection}} &= \frac{1}{|P|} \sum_{j=1}^{|P|} \mathbb{I}\!\left(\frac{S_j}{L_j} \ge \theta\right),
\end{align}

where \( S_i \) is the overlap between true and predicted segments, \( L_i \) is the anomaly duration, and \( \theta \) is the minimum overlap ratio threshold.

\textbf{Positional components:}
\[
R_{\mathrm{position}} = \frac{1}{|A|} \sum_{i=1}^{|A|} \min\left(1, \frac{S_i}{L_i}\right), 
\]
\[
P_{\mathrm{position}} = \frac{1}{|P|} \sum_{j=1}^{|P|} \min\left(1, \frac{S_j}{L_j}\right).
\]

Here, \( S_i = \mathrm{overlap\_score} + \mathrm{decay\_score} \), with the latter computed using a sigmoid decay function over a fuzzy window of length \( \delta \).  

The position dimension uses $\min(1, S_i / L_i)$ to map each anomalous segment into a segment-level score within $[0,1]$, thereby enabling event-level evaluation. However, if the scores of individual segments are subsequently aggregated in a point-weighted manner (e.g., by directly summing $S_i$), long anomaly segments tend to dominate the overall score due to their larger number of points. 
Conversely, if a simple arithmetic mean is taken across all segments, each segment is treated equally, and the importance of longer anomalies is underestimated. 
Therefore, the single definition $\min(1, S_i / L_i)$ is insufficient to balance ``point-level precision'' and ``event-level coverage.'' 
An explicit weighting or hybrid aggregation scheme is required to achieve a proper trade-off between the two.

\paragraph{Enhanced time series aware F-score ($eTaF_\beta$)}
~\\
The enhanced version \( eTaF_\beta \) introduces a length-weighted iterative pruning mechanism to address TaF’s potential over-rewarding of irrelevant overlaps. This refinement yields stricter and more application-aligned evaluation. However, as it inherits the fuzzy zone of length \( \delta \), detailed discussion of its improvement on random-score suppression is deferred to Section~3.5.

\paragraph{Volume Under Surface ($VUS^l$-ROC, $VUS^l$-PR)}
~\\
Traditional AUC-based metrics rely on binary (0/1) labels, which can be overly rigid when anomaly boundaries are uncertain. The \emph{Volume Under Surface (VUS)} metric introduces a soft labeling mechanism with tolerance windows, offering a more robust assessment under label noise and boundary ambiguity~\cite{paparrizos_VolumeSurfaceNew_2022_p@k_VUS}.

\textbf{Computation.} For each true anomaly point, a tolerance window of size \( w \) is defined, and predictions within this window receive distance-weighted scores:
\[
y_t^{(w)} = \max_{i: |i-t| \le w} y_i \cdot f(|i-t|, w),
\]
where the decay function \( f(d, w) \) is typically:
\[
f(d, w) =
\begin{cases}
1 - \frac{d}{w}, & \text{if } d \le w,\\
0, & \text{otherwise.}
\end{cases}
\]
Integrating the AUC values across different tolerance scales yields:
\[
\mathrm{VUS}^l = \int_0^{w_{\max}} \mathrm{AUC}^{(w)} \, dw.
\]
While VUS improves label robustness, it sacrifices the parameter-free property of AUC, making inter-study comparison less direct.

\begingroup

\paragraph{Proximity-Aware Time series Evaluation ($PATE^{\varepsilon\delta}$)}
~\
The label-robust capability of $PATE^{\varepsilon\delta}$ stems from its buffer zones, which constitute 
an explicit modeling of annotation boundary uncertainty.
In practice, anomaly labeling suffers from inherent boundary ambiguity—different annotators may 
disagree on the precise start and end timestamps of the same anomaly by several time steps. 
The pre-buffer ($\varepsilon$) and post-buffer ($\delta$) regions of $PATE^{\varepsilon\delta}$ 
transform hard boundaries into \textbf{soft tolerance intervals}: predictions falling within 
either buffer still receive positive, non-zero weights rather than being categorically 
penalized as false positives or false negatives.

\subsubsection{\textbf{Conceptual Characteristics}}
~\\
Label-robust metrics aim to mitigate annotation uncertainty and temporal boundary ambiguity. 
While III.C.1 defines their formulations, their core differences lie in how ``robustness'' is operationalized and the trade-offs they introduce.

\begin{table}[h]
\begingroup

\centering
\small
\caption{Design mechanisms of label-robust metrics.}
\setlength{\tabcolsep}{4pt}

\begin{tabular}{
p{0.24\columnwidth}
p{0.35\columnwidth}
p{0.35\columnwidth}
}
\toprule
\textbf{Mechanism} & \textbf{Representative Metrics} & \textbf{Primary Cost} \\
\midrule
Tolerance-based relaxation 
& $\mathrm{TTF}_{\beta}^{\tau}$, VUS, $PATE^{\varepsilon\delta}$ 
& Parameter sensitivity (e.g., $d, w, \varepsilon, \delta$) \\

Distance-based quantification 
& TD, AF 
& Reduced emphasis on categorical detection success \\

Decomposition-based evaluation 
& TaF, eTaF 
& Aggregation ambiguity (weighting design) \\
\bottomrule

\end{tabular}
\endgroup

\end{table}

These mechanisms reflect fundamentally different interpretations of robustness. 
Tolerance-based methods relax correctness via explicit buffer regions, improving boundary tolerance but introducing task-dependent hyperparameters. 
Distance-based approaches replace binary matching with continuous proximity modeling, enhancing localization fidelity while weakening discrete detection semantics. 
Decomposition-based designs treat anomaly detection as a multi-dimensional task (e.g., detection, coverage, localization), improving interpretability at the cost of additional aggregation choices.

Overall, label robustness should be understood not as a single property, but as a spectrum of design trade-offs between tolerance flexibility, structural precision, semantic interpretability, and evaluation simplicity.

\endgroup

\subsection{Cost-aware Metrics (Penalizing Human Verification Effort)} 
In real-world anomaly detection systems, evaluation should reflect not only detection accuracy but also the cost of human verification. 
From an operational perspective, different types of false alarms impose unequal burdens on analysts. 
When false positives are densely clustered, multiple alarms can often be verified collectively, leading to relatively low review effort. 
In contrast, dispersed false positives require repeated manual checks, incurring substantially higher verification costs. 
This motivates a class of \emph{sparsity-penalizing metrics} that explicitly punish scattered false alarms to better align with real-world operational costs. 

Meanwhile, another form of verification cost arises from limited analyst capacity. 
In many industrial scenarios, human reviewers can only inspect a small number of high-priority alerts due to time or resource constraints. 
Therefore, rather than global detection coverage, what matters more is how effectively a model prioritizes true anomalies among its top-ranked predictions. 

\subsubsection{\textbf{Formal Definitions}}
~\
\paragraph{Latency and Sparsity-aware F-score (\(\mathit{LSF}^w\))}
~\\
To reduce the verification cost induced by dispersed false alarms, the latency- and sparsity-aware F-score (\(\mathit{LSF}^w\)) integrates the notion of \emph{false alarm resource cost} into the evaluation process~\cite{abdulaal_PracticalApproachAsynchronous_2021_TSAD_latency_sparsity}.

This metric employs a window-based mechanism that simultaneously models both detection delay and false positive sparsity. For false positive aggregation, LSF adopts a \emph{batch-wise} counting scheme: regardless of how many FP points occur within a single time window, they are collectively counted as one FP. This design penalizes dispersed false alarms more heavily than clustered ones, aligning with the real-world observation that scattered false alerts are more disruptive than concentrated bursts.

\begin{figure}
    \centering
    \includegraphics[width=0.55\linewidth]{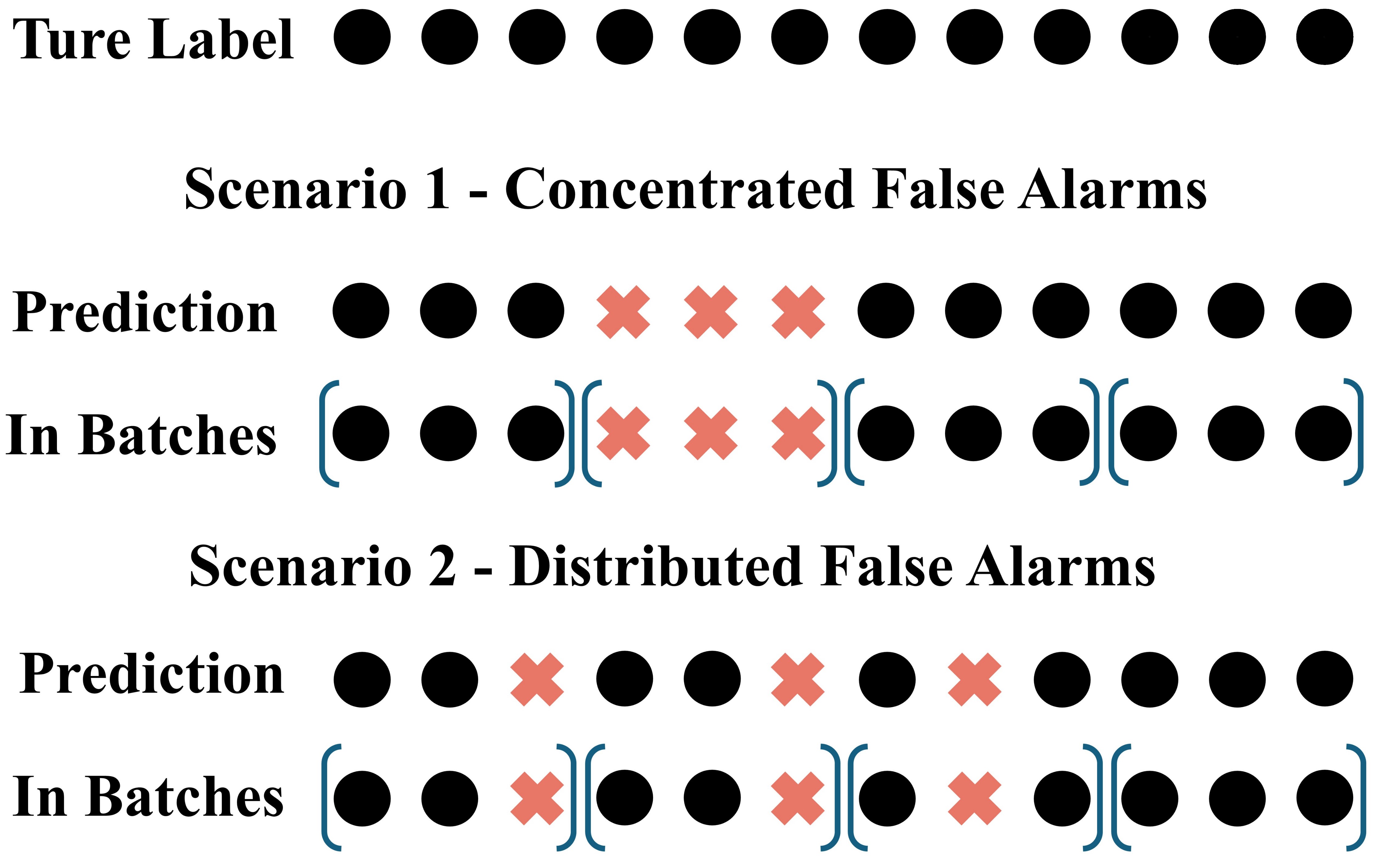}
    \caption{Concentrated vs. Distributed False Alarms under Batch-level Evaluation. Illustration of how \(\mathit{LSF}^w\) differentiates between concentrated and 
distributed false alarms. Although both scenarios contain three point-wise 
false positives, the batch-level penalty differs: only one window is affected 
in Scenario~1, whereas three windows are affected in Scenario~2. This leads to 
a higher penalty for distributed false alarms.}
    \label{fig:placeholder}
\end{figure}

To illustrate how \(\mathit{LSF}^w\) distinguishes between \emph{concentrated} and 
\emph{distributed} false alarms, consider a time series of length 12 with a 
window size of \(w = 3\), as shown in fig~\ref{fig:placeholder}

Under traditional point-wise evaluation, both scenarios yield the same number 
of false positives (\(\mathrm{FP} = 3\)), and thus receive identical penalties:

\textbf{Scenario 1}: false alarms are concentrated in a local region.

\textbf{Scenario 2}: false alarms are scattered across the timeline.

In contrast, \(\mathit{LSF}^w\) evaluates predictions at the 
\textbf{window level}. For each window \(W_i\), the batch-level false positive 
is defined as:
\[
\mathrm{FP}_i = 
\mathbf{1}\!\left[
\max_{t \in W_i} \hat{y}_t = 1 
\ \text{and} \
\max_{t \in W_i} y_t = 0
\right].
\]

This definition implies:If a window contains \emph{any} false positive, the entire window is counted as one batch-level false positive. If a window contains no false positives, it contributes zero.

Therefore, although both scenarios have the same number of point-wise false alarms, their batch-level penalties differ significantly:

\textbf{Scenario 1}: only the second window contains false alarms  
  \(\Rightarrow \mathrm{FP}_{\text{batch}} = 1\).
  
\textbf{Scenario 2}: three windows contain false alarms  
  \(\Rightarrow \mathrm{FP}_{\text{batch}} = 3\).

Consequently, distributed false alarms incur a threefold penalty compared to 
concentrated ones, reflecting their greater operational impact.

\paragraph{NAB Score}
~\\
The \emph{Numenta Anomaly Benchmark} (NAB) score extends traditional accuracy metrics by integrating both timeliness and false positive sparsity~\cite{lavin_EvaluatingRealTimeAnomaly_2015_NAB}. For each true anomaly segment, an equal-length detection window is defined. Predictions within this window receive position-dependent rewards: earlier detections yield higher scores, while delayed detections are penalized progressively. False positives outside the window incur distance-based penalties, which increase with their temporal distance from the nearest true anomaly window. Consequently, concentrated false positives are penalized less than dispersed ones.

While NAB reasonably accounts for human verification cost, its penalty strength for FP sparsity is limited. When false positives occur far away (e.g., beyond three times the window length), their penalties quickly saturate near \(-1\). As a result, a cluster of occasional false positives and an equal number of widely scattered ones receive nearly identical penalties—despite vastly different operational impacts. In real-world review scenarios, the former may require only one manual dismissal, whereas the latter could repeatedly interrupt analyst workflows. Hence, NAB’s treatment of false positive sparsity remains insufficiently rigorous from a cost perspective.

\paragraph{Precision at K (P@K)}
~\\
In many real-world anomaly detection systems, analysts can only review a limited number of high-priority alerts. Thus, rather than global coverage, what matters more is how effectively a model identifies true anomalies among its top-ranked predictions. The Precision@K (\(\mathrm{P@K}\)) metric reflects this concern by evaluating the proportion of true anomalies within the top-\(K\) predicted alerts~\cite{paparrizos_TSBUADEndtoendBenchmark_2022_p@k,paparrizos_VolumeSurfaceNew_2022_p@k_VUS}.

Specifically, anomaly scores \(s_t\) are sorted in descending order, and the top \(K\) predictions are selected as the model’s highest-confidence candidates. The proportion of true anomalies within this subset reflects the model’s effectiveness under constrained review resources. \(\mathrm{P@K}\) directly captures the system’s \emph{alert value density}—whether the model provides sufficiently valuable alerts in the most costly review phase.

\textbf{Computation.} Let \(s_t\) denote the anomaly scores and \(y_t\) the ground-truth labels. Then:
\[
K = \sum_{t=1}^{T} y_t, \quad
\tau = \mathrm{sort}(s_t)_{[-K]}, \]
\[
\hat{y}_t = \mathbb{I}[s_t \ge \tau], \]
\[
P@K = \frac{\sum_{t=1}^{T} \hat{y}_t \cdot y_t}{\sum_{t=1}^{T} \hat{y}_t},
\]
where \(\mathbb{I}[\cdot]\) is the indicator function, and \(\mathrm{sort}(s_t)_{[-K]}\) denotes the \(K\)-th largest score.

\begingroup

\subsubsection{\textbf{Conceptual Characteristics}}
~\\
Cost-aware metrics extend traditional accuracy-based evaluation by explicitly modeling human verification effort as part of the objective. Rather than treating all false positives equally, they reinterpret anomaly detection performance through the lens of operational burden.

Despite sharing this cost-oriented motivation, the metrics in this category encode verification effort in fundamentally different ways:

\paragraph{Sparsity-Sensitive Penalization}

Metrics such as \textit{LSFw} and \textit{LSF}, and to a lesser extent NAB, model verification cost through the temporal distribution of false positives.

Their central assumption is: The operational burden depends not only on how many false alarms occur, but also on how they are distributed in time.

Two modeling strategies emerge:

\begin{itemize}
    \item \textbf{Window-level aggregation (LSF)}\\
    False positives within a time window are treated as a single verification batch. This converts point-wise errors into event-level cost units, directly penalizing dispersion.

    \item \textbf{Distance-based reward/penalty shaping (NAB)}\\
    Timeliness and sparsity are incorporated via position-dependent reward curves. False positives far from anomaly windows incur penalties, but the cost function eventually saturates.
\end{itemize}

The key distinction lies in cost granularity:

\begin{itemize}
    \item LSF discretizes review effort into batch-level verification units.
    \item NAB applies a continuous temporal reward function but does not strictly differentiate highly dispersed alarms once penalties saturate.
\end{itemize}

Thus, sparsity-aware metrics operationalize cost through temporal clustering structure, redefining false positives as workload events rather than isolated points.

\paragraph{Budget-Constrained Prioritization}

In contrast, \textit{P@K} models a different type of cost: limited analyst capacity.

Here, evaluation shifts from “How many anomalies are detected?” to 
“Are the most important alerts surfaced first?”

Under this perspective, only the top-ranked predictions are economically relevant, 
and detections outside the review budget are operationally inconsequential.

The metric becomes inherently ranking-centric. Unlike sparsity-based metrics, 
which reshape penalties in the time domain, \textit{P@K} operates in the score-ranking domain. 
It does not modify error definitions but restricts evaluation to a capacity-constrained subset.

This transforms anomaly detection into a prioritization problem rather than a global coverage problem. 
Importantly, this paradigm is fundamentally different from sparsity-sensitive cost modeling: 
the former assumes a strictly capped review budget, while the latter models the variable cost of repeated verification. 
As a result, the two are not interchangeable, and optimizing for one may degrade performance under the other.
\endgroup

\subsection{Robustness to Random or Sparse Scoring}

A critical yet often overlooked issue in evaluating time series anomaly detection lies in the vulnerability of certain metrics to random or sparse prediction behaviors. 
\cite{kim_RigorousEvaluationTimeSeries_2022_k_Point_adjusted} reported that, under specific configurations, random guessing could yield higher scores than state-of-the-art models when using the conventional point-adjusted F1 metric($PAF_{\beta}$). 
This phenomenon occurs because the metric treats an entire anomalous segment as correctly detected once any single point within it is identified, regardless of the overall detection quality. 
As a result, sparse or randomly scattered detections may undeservedly achieve inflated scores, undermining the credibility of evaluation results.

To mitigate this, several subsequent studies have introduced mechanisms that explicitly resist random or overly sparse scoring—for example, by imposing temporal consistency requirements or rewarding detection coverage rather than isolated hits. 
Nevertheless, not all robustness stems from explicit design: 
some metrics inherently resist random scoring due to their point-based nature, which evaluates detections at a finer temporal granularity. 

\begingroup

In this classification subsection, some metrics inherently exhibit robustness to random or sparse scoring due to their design. Therefore, metrics are associated with this dimension primarily based on their design principles rather than solely on empirical performance. This design-oriented classification is followed by a systematic empirical analysis in Section~\ref{sec:randomness_experiment}, which examines how these metrics behave under random or uninformed predictions and identifies those that fail to distinguish genuine detector outputs from such baselines.

\endgroup

\subsubsection{\textbf{Formal Definitions}}
~\
\paragraph{Time series aware F-score (\(\mathit{TaF}_{\beta}^{\delta}\))}
~\\
\emph{\(\mathit{TaF}_{\beta}^{\delta}\)}~\cite{hwang_TimeSeriesAwarePrecision_2019_tims_serise_aware_taf} introduces a \emph{coverage threshold} to ensure that model predictions have substantive temporal overlap with true anomaly segments, thereby mitigating artificially high scores caused by random or sparse predictions. As shown in Eq.~\ref{TaF_1}, the computation is performed at the \emph{event level}. The coverage threshold \(\theta\) requires that a prediction must cover at least a certain fraction of a true anomaly segment to be considered a successful detection; otherwise, the entire segment is treated as a miss. This prevents models from achieving inflated scores merely through “spot hits”.

\paragraph{Enhanced time series aware F-score (\(\mathit{eTaF}_{\beta}\))}
~\\
As noted by Hwang et al.~\cite{hwang_YouKnowExisting_2022_Enhanced_time_aware_etaf}, although TaF alleviates the inflation caused by random hits, it may still reward ``overlapping but meaningless'' detections, where low-quality predictions can still contribute substantial scores. Moreover, TaF lacks a proper length-weighting mechanism, which causes inconsistencies between event- and point-level contributions. To address these limitations, the  \emph{Enhanced time series aware F-score} (\(\mathit{eTaF}_{\beta}\)) introduces an \emph{iterative pruning mechanism} and \emph{length-weighted scoring}, making the evaluation stricter and more realistic.

The core principle of eTaF is that a valid detection must not only overlap sufficiently with the true anomaly segment but also avoid excessive overextension or under-coverage. In implementation, eTaF first constructs an overlap matrix between true and predicted segments, then performs iterative pruning to remove low-quality matches: true segments with recall below \(\theta_r\) and predicted segments with precision below \(\theta_p\) are discarded. This process is repeated until convergence.

After pruning, the retained segments are evaluated by combining detection judgment, overlap ratio, and length-weighted contribution.  
For instance, each predicted segment is weighted by the square root of its length to prevent disproportionately large contributions from long segments.  
The resulting eTaR and eTaP represent the recall and precision components, respectively:

\[
\mathrm{eTaR} = \frac{1}{2|A|}\sum_{i=1}^{|A|} (d_i + d_i p_i), \quad
\mathrm{eTaP} = \frac{1}{2W}\sum_{j=1}^{|P|} \sqrt{L_j} (d_j + d_j p_j),
\]
where \(d_i\) denotes the detection score, \(p_i\) the localization score, and \(W\) the normalization constant.  
The thresholds \(\theta_r\) and \(\theta_p\) define the minimum acceptable recall and precision levels, while \(\delta\) controls the fuzzy boundary proportion.

Compared to TaF, eTaF offers several advantages:  
(1) the pruning mechanism effectively filters out low-quality matches, reducing inflated results;  
(2) the dual-threshold constraints guarantee minimal performance standards for both recall and precision; and  
(3) length weighting better balances the influence of segments of varying durations.  
However, this strictness comes at a cost—iterative pruning may remove numerous matches, and tuning \(\theta_r\) and \(\theta_p\) requires task-specific calibration. Furthermore, eTaF incurs higher computational complexity than basic metrics.

\paragraph{K\%-Point-adjusted F-score (\(\mathit{K\%\text{-}PAF^k_{\beta}}\))}
~\\
In some anomaly detection tasks, the key concern is not merely whether an anomaly is detected, but whether the detected segment sufficiently \emph{covers} the anomalous process.  
For instance, in medical monitoring or intrusion detection, isolated and short-lived alarms often lack explanatory value, whereas sustained detections are far more informative.  
To capture this intuition, the \emph{K\%-Point-adjusted F-score} (\(\mathit{K\%\!-\!PAF}_{\beta}\)) enforces a minimum coverage constraint: only if a predicted segment covers at least \(K\%\) of the true anomaly points is it considered a successful detection~\cite{kim_RigorousEvaluationTimeSeries_2022_k_Point_adjusted,hwang2022you_k_PAF,hwang2019time_k_PAF}. This constraint prevents models from achieving high scores through accidental or random “flashing” detections.

\textbf{Computation.} Let the true anomaly segment be \([s, e]\) with length \(L = e - s + 1\).  
If the number of predicted anomaly points within this segment satisfies \(\mathrm{hit} \ge K \cdot L\), the entire segment is marked as detected:
\[
y_t = 1, \quad \forall t \in [s, e].
\]

Otherwise, the predicted points within the segment are removed.  
The final F-score is then computed based on the adjusted labels:
\[
\mathit{K\%\text{-}PAF^k_{\beta}} = \frac{(1 + \beta^2) \cdot \mathrm{TP}}{(1 + \beta^2) \cdot \mathrm{TP} + \beta^2 \cdot \mathrm{FN} + \mathrm{FP}}.
\]

\begingroup

\subsubsection{\textbf{Conceptual Characteristics}}
~\\
Robustness to random or sparse scoring addresses a fundamental evaluation question:

\textit{Can a metric reliably distinguish meaningful detection behavior from accidental or uninformed predictions?}

Unlike label-robust or cost-aware metrics, which primarily modify tolerance or operational weighting, this category focuses on preventing artificial score inflation caused by isolated hits or fragmented detections. Although the representative metrics share this objective, they operationalize robustness through distinct structural mechanisms.

\paragraph{Coverage-Constrained Event Validation}

Metrics such as \(K\%\text{-PAF}_\beta\), \(K\%\text{-PAF}\), and partially \( \text{TaF}_{\beta}^{\delta} \) introduce minimum coverage requirements at the event level. Core principle is: An anomaly segment is considered detected only if a sufficient portion of it is captured.

This shifts evaluation from the existence of any overlap to the \emph{adequacy} of overlap. Under traditional point-adjusted metrics, a single predicted point can validate an entire anomaly segment. Coverage-constrained metrics reject this binary shortcut by enforcing a quantitative threshold.

Structurally, this mechanism transforms detection validation into a segment-level completeness test, suppressing random or sporadic hits that lack temporal consistency.

However, this strictness introduces sensitivity to segment length and threshold selection. When anomalies are intrinsically short, coverage constraints may underestimate practically acceptable detections.

\paragraph{Iterative Quality Filtering and Dual Constraints.}

The enhanced \( \text{eTaF}_\beta \) adopts a more rigorous strategy by combining: minimum recall thresholds, minimum precision thresholds, iterative pruning of low-quality matches. Rather than applying a single coverage condition, it performs mutual validation between predicted and true segments. This design reframes evaluation as a constrained matching process:

\begin{itemize}
    \item overlapping segments must satisfy both recall and precision criteria,
    \item matches failing to meet quality standards are progressively removed,
    \item final scores reflect only structurally consistent detections.
\end{itemize}

Compared to simple coverage thresholds, this approach provides stronger resistance to random scoring because isolated or noisy matches are unlikely to survive iterative filtering. The trade-off lies in increased complexity and parameter dependence. Robustness here is achieved through stricter structural admissibility rather than relaxed tolerance.

\paragraph{Granularity as Implicit Robustness.}

Beyond explicit constraints, robustness may also emerge implicitly from evaluation granularity.

\begin{itemize}
    \item \textbf{Point-level metrics} inherently penalize sparse predictions because each predicted point contributes independently to precision and recall.
    \item \textbf{Event-adjusted metrics} are more vulnerable to random inflation if they treat entire segments as atomic units.
\end{itemize}

Thus, robustness is not solely a matter of adding constraints; it is also determined by the fundamental granularity at which correctness is defined. Fine-grained evaluation reduces the probability that accidental hits disproportionately influence global performance.

\paragraph{Structural Trade-offs.} From a broader perspective, robustness-enhancing metrics balance two competing goals: A. avoiding inflated scores from random or sparse predictions. B. preserving sensitivity to partial but meaningful detections. A metric’s ability to resist random inflation ultimately reflects how rigorously it defines what constitutes a \emph{valid detection}.

\endgroup

\subsection{Threshold-independent and Parameter-free Metrics}

In anomaly detection, continuous anomaly scores must be converted into binary 
predictions via a threshold to compute conventional classification metrics. 
However, the selection of such thresholds is often subjective and 
data-dependent.
So, this subsection introduces evaluation metrics that eliminate explicit threshold dependence or avoid user-defined hyperparameters. Their common characteristic is that performance is assessed either across all possible thresholds or directly from intrinsic data structure without additional tuning.

\subsubsection{\textbf{Formal Definitions}}
~\
\paragraph{Best Threshold Binary Metrics ($\mathit{Best\text{-}PwF}$)}
~\\
This method exhaustively searches over all possible threshold candidates. For each threshold, a binary classification performance metric (e.g., the F$_1$ score) is computed, and the maximum value among them is taken as the final evaluation result~\cite{liu_TimeSeriesAnomaly_2023,liu2022time_Best-PwF,deng2021graph_Best-PwF,huang2022semi_Best-PwF,lavin2015evaluating_Best-PwF,campos2021unsupervised_Best-PwF}. Essentially, this approach represents the performance upper limit of a model under \emph{perfect threshold tuning}.

\textbf{Computation.} Let the anomaly score sequence be $\{s_t\}_{t=1}^T$ and the predicted binary labels be $\{\hat{y}_t\}_{t=1}^T$.  
The F$_1$ score under the best threshold is defined as:
\[
Best\text{-}PwF = \max_{\theta \in s_t} 
F_1\!\left(\hat{y}_t = \mathbb{I}[s_t \ge \theta]\right),
\]
where:
\begin{itemize}
  \item $\theta$ iterates over all unique values in $\{s_t\}$,
  \item $\mathbb{I}[\cdot]$ is the indicator function,
  \item $F_1(\cdot)$ is the standard F$_1$-score computation function.
\end{itemize}

\paragraph{Area Under Curve (AUC-ROC, AUC-PR)}
~\\
Traditional binary metrics rely on a fixed threshold, which introduces subjectivity and makes optimal threshold selection difficult in practice.  
To achieve a more comprehensive and objective evaluation, area-under-curve (AUC) metrics assess model performance across all possible thresholds, thereby eliminating the dependence on any single threshold choice~\cite{feng_TimeSeriesAnomaly_2021_AUC,bhatia_MStreamFastAnomaly_2021_AUC,kieu_OutlierDetectionTime_2019_AUC,park_MultimodalAnomalyDetector_2018_AUC,dai_GraphaugmentedNormalizingFlows_2022_AUC,ergen_UnsupervisedAnomalyDetection_2020_AUC,goodge_RobustnessAutoencodersAnomaly_2020_AUC,li2021dct_AUC,lobo2008auc_AUC,dai2022graph_AUC,dai2021sdfvae_AUC,wang2019study_AUC,zhang2019velc_AUC,kieu2019outlier_AUC,baker2001proposed_AUC,berrar2012caveats_AUC,bhatia2021mstream_AUC,schmidl2022anomaly_nab,huang2020crowdquake_AUC,feng2022unsupervised_AUC,goodge2021robustness_AUC,davis2006relationship_AUC,ergen2019unsupervised_AUC,campos2021unsupervised_AUC}.

The key idea behind AUC-based metrics is to quantify a model’s overall discriminative ability by integrating performance over varying thresholds.  
Specifically, \emph{AUC-ROC} (Receiver Operating Characteristic Curve) uses the True Positive Rate (TPR, i.e., recall) as the vertical axis and the False Positive Rate (FPR) as the horizontal axis, capturing the classifier’s global discrimination power between positive and negative samples.  
In contrast, \emph{AUC-PR} (Precision–Recall Curve) plots Precision against Recall, focusing more on the model’s ability to detect positive instances. This makes AUC-PR particularly suitable for highly imbalanced anomaly detection tasks.

\textbf{Computation.} Let $s_t$ denote the anomaly score and $y_t$ the ground-truth label. Then:
\[
\mathrm{AUC\text{-}ROC} = \int_{0}^{1} \mathrm{TPR}(\tau)\, d(\mathrm{FPR}(\tau)), \]
\[
\mathrm{AUC\text{-}PR} = \int_{0}^{1} \mathrm{Precision}(\tau)\, d(\mathrm{Recall}(\tau)),
\]
where $\tau$ denotes the threshold parameter traversing all possible settings.

From these definitions:
\begin{itemize}
  \item \textbf{AUC-ROC} is suitable for balanced-class scenarios or when global discrimination capability is of interest.
  \item \textbf{AUC-PR} is more appropriate for anomaly or rare-event detection tasks, where the positive class is scarce.
\end{itemize}

\paragraph{Temporal distance (TD)}
~\\
TD addresses the issue of annotation uncertainty, but its design is inherently parameter-free, relying entirely on the intrinsic structure of the data rather than any manually specified hyperparameters. As shown in Eq.~\ref{TD}, TD computes a symmetric bidirectional nearest-neighbor distance between predicted and ground-truth anomalous point sets, thereby directly measuring the temporal misalignment between them.

TD reformulates anomaly detection assessment as a point-set matching problem on the time axis, where the evaluation is performed through a deterministic minimum-distance aggregation. The entire procedure contains no tunable components, and every essential element is naturally determined by the input data itself. This design ensures that TD maintains a consistent evaluation standard across datasets and application scenarios, eliminating the potential influence of parameter selection on the evaluation outcome.

\paragraph{Affiliation-based F-score ($AF_\beta$)}
~\\
As previously discussed, the integral form of affiliation metrics naturally 
provides robustness to boundary ambiguity, and its core mechanism is also fully 
data-driven without relying on any external hyperparameter specification. 
Excluding the weighting parameter in the F-score, when $\beta = 1$ (i.e., the 
standard $F1$ setting), the AF metric becomes a purely data-driven evaluation 
method that requires no parameter tuning for its application. No manually defined tolerance regions or buffer windows are required. The parameter-free nature of $AF_\beta$ originates from the geometric essence of its “affiliation relation”: it partitions the time axis into a set of non-overlapping responsibility regions and evaluates the local prediction fidelity through distance-based integration within each region.

This formulation avoids the binary “perfect match vs. complete mismatch’’ dilemma present in traditional metrics, replacing it with a continuous probabilistic scoring mechanism that smoothly reflects subtle variations in prediction quality. More importantly, the boundaries of the affiliation regions $E_i$ are automatically determined by the midpoints between neighboring ground-truth anomalous segments. This automatic region construction guarantees cross-dataset consistency—regardless of the number, length, or spatial distribution of anomalous segments, $AF_\beta$ evaluates predictions under a unified logical framework without requiring parameter adjustments for different scenarios.

\begingroup

\subsubsection{\textbf{Conceptual Characteristics}}
~\\
Although all metrics in this category avoid explicit user-specified thresholds during evaluation, their conceptual motivations differ.

Best-PwF represents a theoretical upper bound under perfect threshold tuning.

AUC-based metrics assess ranking quality by integrating performance over all possible decision thresholds, thereby avoiding reliance on any single operating point.

TD reframes anomaly detection as a geometric alignment problem on the time axis.

AF replaces binary matching with continuous affiliation-based integration.

Two distinct philosophies emerge:

\begin{itemize}
 \item  Optimization-based threshold removal (Best-PwF, AUC):
Threshold dependence is mitigated by evaluating over the entire threshold space.

 \item  Structure-driven evaluation (TD, AF):
Evaluation is derived directly from intrinsic temporal structure without external hyperparameters.
\end{itemize}

These approaches eliminate manual tuning, but preserve fundamentally different interpretations of detection quality.
\endgroup

For completeness, we note that the NAB score is a parameterized metric by 
design, yet all of its hyperparameters have been standardized and fixed in the 
official specification. Consequently, it requires no additional parameter 
configuration during use. Since these preset values do not introduce any 
evaluation mechanisms beyond the scope of this discussion, we do not further 
elaborate on NAB here.

\section{Experiments}
\label{sec:randomness_experiment}

\begingroup

This section evaluates the robustness, or ``antirandomness,'' of existing time series anomaly detection metrics when confronted with random or uninformed predictions.
All experiments are conducted under strictly controlled and reproducible settings, encompassing both synthetic datasets with diverse anomaly configurations and \emph{real-world industrial benchmarks} drawn from TimeSeriesBench~\cite{si2024timeseriesbench}, along with various random-prediction strategies and multiple tiers of detector quality (referred to as the \emph{quality gradient}).
Each metric is comprehensively assessed through a suite of statistical indicators to examine the following hypotheses:
\begin{enumerate}
    \item Genuine detectors should yield significantly higher metric scores than random predictions;
    \item Metrics should demonstrate monotonic consistency along the detection-quality gradient — as detector performance improves (from random to genuine), the corresponding metric scores are expected to increase accordingly.
\end{enumerate}

\endgroup

\subsection{Datasets and Synthetic Configuration}

This experiment systematically evaluates the stability and robustness of different anomaly detection metrics under ``uninformative'' or ``misleading'' prediction scenarios.
The evaluation data are drawn from two complementary sources: \emph{synthetic datasets} with precisely controlled anomaly properties, and \emph{real-world industrial datasets} from established benchmarks.

\begingroup

\paragraph{Real-World Datasets}

To validate that the conclusions drawn from synthetic data generalize to practical scenarios, three widely-used univariate time series (UTS) anomaly detection benchmarks are incorporated via the EasyTSAD framework, all originating from TimeSeriesBench:

\begin{itemize}
    \item \textbf{AIOPS} — Operational metrics from large-scale internet services, featuring diverse failure modes including traffic spikes, latency anomalies, and resource saturation. Due to the extreme length of certain AIOPS sequences, a truncation ratio of 15\% is applied to prevent prohibitive computation time for threshold-sensitive metrics.
    \item \textbf{NAB} — The Numenta Anomaly Benchmark, containing streaming time series from domains such as IT infrastructure, social media, and transportation, with both point and collective anomalies.
    \item \textbf{Yahoo} — The Yahoo S5 benchmark (Webscope), consisting of real and synthetic server metrics with labeled anomalies covering level shifts, spikes, and trend changes.
\end{itemize}

For each benchmark, all series are preprocessed with z-score normalization.

\endgroup

\paragraph{Synthetic Data Generation}

The synthetic data generation process follows two principles: \emph{coverage} and \emph{specificity}.

Coverage ensures that the constructed data encompass diverse temporal structures and statistical properties, allowing for a comprehensive examination of metric generality across distributions.
Specificity, in contrast, introduces extreme or fragile cases (e.g., isolated anomalies, long-term drifts, periodic perturbations) to assess potential failure risks at distributional boundaries.

It is important to emphasize that this study focuses on the \emph{validation of metrics} rather than the absolute performance of detection models.
Therefore, certain scenarios are deliberately designed to yield deceptively high metric scores, thereby revealing possible risks of misleading evaluation in practical applications.

\paragraph{Anomaly Types and Generation Mechanisms}

The synthetic data include five representative anomaly types to ensure comprehensive coverage across different semantic categories. Their proportions, generation mechanisms, and parameter ranges are summarized below:

\begingroup

\begin{itemize}
    \item \textbf{Point anomalies (2.5\%)} — Random isolated spikes simulating transient faults or sensor glitches.
    \item \textbf{Level shifts (35\%)} — Upward or downward shifts of varying length, representing short- to long-term mean or amplitude drifts.
    \item \textbf{Collective anomalies (25\%)} — Persistent segment-level behavioral changes, including frequency changes, noise amplification, and pattern alterations.
    \item \textbf{Periodic disruption (25\%)} — Local disruptions of periodic structure to test metric adaptability in periodic contexts.
    \item \textbf{Contextual anomalies (12.5\%)} — Locally significant deviations within seasonal or trending contexts, challenging the distinction of local vs. global anomalies.
\end{itemize}

\endgroup
To ensure consistent statistical characteristics, the baseline standard deviation $\sigma_0$ is computed from the clean signal before anomaly injection.
All anomaly intensities are defined relative to $\sigma_0$, preventing uncontrolled amplification effects.
Random seeds and diverse parameter sampling strategies are employed to guarantee robustness and reproducibility of results.

\paragraph{Scale and Sparsity Design}

To systematically assess metric behavior under varying data scales and anomaly densities, a stratified sampling design is adopted with three control dimensions: sequence length, contamination rate, and repetition.

Sequence lengths are set to 5k, 10k, and 50k, corresponding to short, medium, and long time series scenarios.
Anomaly segment lengths and local window parameters are scaled proportionally to maintain comparability.
Contamination rates are set to 5\%, 10\%, 15\%, and 20\%, covering sparse, moderate, and dense anomaly distributions.
For each synthetic configuration, 9 independent datasets are generated with distinct random seeds to control stochastic variation. Combined with the real-world benchmarks (up to 3 curves $\times$ 3 datasets), this yields a diverse evaluation corpus spanning both controlled and naturalistic conditions.

\subsection{Prediction Strategies and Control Groups}

This section introduces the prediction and control strategies used to validate the robustness of metrics.
The core idea is to construct a continuous degradation sequence from high-quality predictions to purely random ones (termed the \emph{quality gradient}), enabling systematic comparison of metric responses under both informative and spurious detection conditions.

This design tests two essential properties:
\begin{enumerate}
    \item whether a metric correctly reflects detection quality differences (monotonicity);
    \item whether it can effectively distinguish informative predictions from random noise (anti-randomness).
\end{enumerate}

\begingroup

\paragraph{Genuine Detector Group}

To represent ``informative detectors,'' four representative anomaly detection models are selected from the EasyTSAD framework, covering distinct modeling paradigms:
\begin{itemize}
    \item \emph{AR} (Auto-Regressive) — a classical statistical model based on linear autoregression, serving as the baseline for traditional time series analysis;
    \item \emph{LOF} (Local Outlier Factor) — a density-based machine learning method that identifies anomalies via local density deviation, representing unsupervised outlier detection;
    \item \emph{TimesNet} — a deep learning model based on temporal 2D-variation modeling with inception-style architecture, representing the prediction-based paradigm;
    \item \emph{AE} (Autoencoder) — a reconstruction-based deep learning model that detects anomalies through reconstruction error, representing the reconstruction paradigm.
\end{itemize}

These four models span a broad spectrum of detection methodologies — from classical statistics and machine learning to deep prediction and deep reconstruction — ensuring that the genuine detector group adequately represents the diversity of real-world detection pipelines.

\endgroup

\paragraph{Quality Gradient Mechanism}

To examine sensitivity and monotonic response, a \emph{Quality Gradient} mechanism is designed to generate progressively degraded predictions.
Controlled perturbations gradually reduce the alignment between predictions and ground truth, yielding a sequence of prediction qualities $\alpha \in \{0.9, 0.8, \ldots, 0.1\}$.

Prediction degradation is controlled by flipping ratios and noise amplitude:
\begin{itemize}
    \item high-quality region ($\alpha > 0.7$): mild perturbation — ground truth labels with a small fraction of flipped positions and low-variance Gaussian noise ($\sigma = 0.1$);
    \item medium-quality region ($0.4 < \alpha \le 0.7$): moderate noise and flipping — increased flip ratio and higher noise ($\sigma = 0.2$);
    \item low-quality region ($\alpha \le 0.4$): transition to random Bernoulli predictions with uniform scores, becoming completely uninformative.
\end{itemize}

The monotonicity of each metric is quantified by the Spearman rank correlation coefficient $\rho$ between metric scores and quality levels.
A significantly positive $\rho$ close to 1 indicates stable monotonic behavior, whereas $\rho \approx 0$ or negative implies poor quality sensitivity.

\paragraph{Random Prediction Groups (Control Baselines)}

To systematically test metric resistance against uninformative or pseudo-structured predictions, 
we design three random prediction strategies that cover distinct types of randomness and attack scenarios. 
Each strategy is executed 20 times to obtain empirical score distributions, providing robust baselines 
for evaluating metric discrimination power.

\begin{itemize}
    \item \textbf{Uniform Score Random} — This baseline represents purely random scores without any structure. 
          Scores are independently sampled from $\mathcal{U}(0,1)$. To match the empirical anomaly rate $p = \mathbb{E}[y_t]$
          in the ground truth, we apply percentile-based thresholding: the $k = \lfloor p \cdot n \rfloor$-th largest scores 
          are selected as alarms, where $n$ is the sequence length. This ensures the random predictions maintain the same 
          expected anomaly rate as real anomalies, creating a conservative baseline that forces metrics to distinguish 
          \emph{quality} (not just prevalence) of anomaly detection.
          
    \item \textbf{Clustered Random} — This baseline simulates spurious structural patterns or noise bursts that might 
          accidentally correlate with anomalies. Alarms are generated via a cluster-based process:
          \begin{enumerate}
              \item Initialize all scores to $\mathcal{U}(0, 0.3)$ (low baseline).
              \item Generate $m = \max(1, \lfloor p \cdot n / 10 \rfloor)$ clusters (clusters are $\approx 10\times$ sparser than anomalies).
              \item For each cluster: randomly select a start position, sample cluster size from $[5, 15]$ points, 
                    and set scores within the cluster to $\mathcal{U}(0.6, 1.0)$ (high).
              \item Apply threshold $\tau = 0.5$ to convert scores to binary predictions.
          \end{enumerate}
          Unlike uniform random, clustered random introduces spatial correlation and structural bias, testing whether metrics 
          are fooled by incidental bursts of alarms that lack causal relationship with true anomalies.
          
    \item \textbf{Pure Bernoulli Random} — This baseline represents fully independent, uncorrelated randomness. 
          Binary predictions are independently sampled as $a_t \sim \mathrm{Bernoulli}(p)$ with no spatial or temporal structure. 
          Confidence scores are generated as $s_t = a_t + \mathcal{N}(0, 0.2)$ and clipped to $[0,1]$, providing continuous 
          score outputs that correlate imperfectly with binary predictions. This baseline is the most conservative (hardest for 
          metrics to pass) because it provides no structure whatsoever—if a metric fails against Bernoulli random, 
          it is highly vulnerable to adversarial attacks.
\end{itemize}

\noindent\textbf{Rationale:} These three strategies cover a spectrum of randomness:
\begin{itemize}
    \item Uniform tests whether metrics can distinguish informed patterns from noise with preserved prevalence.
    \item Clustered tests whether metrics are robust to accidental structural correlations (e.g., data artifacts, seasonal bursts).
    \item Bernoulli tests the absolute discrimination power against purely uncorrelated randomness.
\end{itemize}
\begingroup

\paragraph{Oracle Attack (Gaming Resistance)}

To further assess \emph{gaming resistance}, an \textbf{Oracle Attack} mechanism is introduced, simulating an adversarial detector that exploits ground truth knowledge to maximize the target metric. Three complementary attack strategies are employed, generating a total of five adversarial predictions per dataset:

\begin{enumerate}
    \item \textbf{Oracle Hill-Climbing} ($\times 1$) — A greedy search that initializes a random alarm set under an alarm-rate constraint $\|\hat{y}\|_1 \le 0.05 L$ and iteratively moves or flips alarm positions to maximize pointwise $F_1$, running for 500 iterations.

    \item \textbf{Boundary-Aware} ($\times 3$) — For each ground-truth anomaly segment, alarms are concentrated within a narrow window ($\pm 2$--$5$ points) around the segment boundaries. Three bias levels $b \in \{0.5, 0.7, 0.9\}$ control the probability of placing alarms at each boundary, producing predictions that partially overlap with true events but provide minimal interior coverage. This specifically targets metrics that are lenient toward boundary-adjacent detections.

    \item \textbf{Overlap-Aware} ($\times 1$) — Within each ground-truth segment, a small number of short alarm windows (length $\approx 3$) are placed with 30\% probability per candidate position, creating sparse partial overlaps. This targets metrics that reward any overlap with true segments regardless of coverage quality.
\end{enumerate}

\endgroup

For all adversarial predictions, anomaly scores are assigned as:
\[
s_t \sim
\begin{cases}
U(0.7, 1.0), & \text{if } \hat{y}_t = 1,\\
U(0.0, 0.3), & \text{otherwise,}
\end{cases}
\]
ensuring that scores are consistent with labels. If a metric can be easily inflated through such manipulation, its evaluation reliability is questionable.

\begingroup

\subsection{Sensitivity to Dataset Characteristics}

To examine whether the discriminative ability of evaluation metrics remains stable across datasets with different properties, we conduct a sensitivity analysis with respect to three dataset-level characteristics: \emph{anomaly density} (contamination rate), \emph{time-series length}, and \emph{signal-to-noise ratio} (SNR).

Since the original time-series noise level is not consistently available across benchmark datasets, we use a score-based proxy for detection difficulty. Specifically, the SNR is defined as

\[
\text{SNR} = 
\frac{\mu_{\text{anomaly scores}} - \mu_{\text{normal scores}}}
{\sigma_{\text{normal scores}}},
\]

where $\mu_{\text{anomaly scores}}$ and $\mu_{\text{normal scores}}$ denote the mean anomaly scores on anomalous and normal regions, respectively, and $\sigma_{\text{normal scores}}$ is the standard deviation of scores in normal regions. Intuitively, a lower SNR indicates that anomaly scores are less distinguishable from normal scores, corresponding to a more challenging detection scenario.

For each (metric, feature) pair, we compute the Spearman rank correlation coefficient $\phi$ between the dataset feature values and the metric's separation performance across all dataset--curve instances. Two separation indicators are considered: \emph{Effect Size} and \emph{AUC-based separation}. 

The coefficient $\phi$ measures the monotonic dependence between dataset characteristics and the metric's discriminative ability. A value close to zero ($|\phi|\approx0$) suggests that the metric’s separation capability is largely insensitive to the corresponding dataset property, whereas a large $|\phi|$ with statistical significance ($p<0.05$) indicates that the metric’s discriminative performance may vary systematically under different data conditions.

In addition to dataset-level characteristics, we further investigate whether the choice of anomaly detection model influences the evaluation conclusions. Four representative detectors spanning a broad complexity spectrum are considered: AR (statistical baseline), LOF (classical machine learning), TimesNet (deep forecasting), and AE (deep reconstruction). For each model--dataset pair, we compute every metric on the genuine model predictions and measure the separation from random-prediction baselines using the $z$-score:
\[
z = \frac{s_{\text{genuine}} - \mu_{\text{random}}}{\sigma_{\text{random}}},
\]
where $s_{\text{genuine}}$ is the metric score of a real detector and $\mu_{\text{random}}$, $\sigma_{\text{random}}$ are the mean and standard deviation of the same metric evaluated on random predictions. A positive $z$-score indicates that the metric successfully distinguishes the genuine detector from random noise; larger values correspond to stronger separation. To prevent a small number of dataset--metric pairs with near-zero $\sigma_{\text{random}}$ from dominating the visualization, $z$-scores are winsorized to $[-20, 20]$.

\endgroup

\subsection{Evaluation Metrics}

The goal of this experiment is not to assess detector performance but to verify whether evaluation metrics can effectively distinguish informative detectors from random guessing. 
Three complementary quantitative dimensions are used: 
\emph{Average Cohen’s effect size $d$}, \emph{Average AUC}, and \emph{Monotonicity coefficient ($\rho$)}. 
These collectively form a comprehensive framework to assess \emph{discriminability}, \emph{robustness}, and \emph{interpretability}.

\paragraph{Average Cohen’s effect size $d$}
To quantify the sensitivity of each metric to score differences between genuine detectors and random predictors, we compute the standardized mean difference:
\[
d = \frac{\mu_{\mathrm{genuine}} - \mu_{\mathrm{random}}}{\sigma_{\mathrm{pooled}}},
\]
where $\mu_{\mathrm{genuine}}$ and $\mu_{\mathrm{random}}$ denote the mean scores of genuine and random predictors, respectively, and $\sigma_{\mathrm{pooled}}$ is their pooled standard deviation. 
Larger $d$ values indicate stronger discriminative capability.

\paragraph{Average AUC}
To further evaluate global separability, we treat metric scores as decision functions and compute the area under the ROC curve (AUC), 
where random samples are labeled as 0 (negative) and genuine detectors as 1 (positive). 
AUC values range from $0.5$ (no discrimination) to $1.0$ (perfect discrimination).

\paragraph{Monotonicity Coefficient ($\rho$)}
To assess the consistency of metric response to detection quality, the Spearman rank correlation coefficient $\rho$ between metric scores and quality levels is computed. 
Following empirical thresholds, $\rho \ge 0.8$ indicates strong monotonicity, whereas $\rho < 0.2$ suggests no significant monotonic relationship.

\begin{table}[!htbp]
\begingroup

\centering
\scriptsize
\setlength{\tabcolsep}{3pt} 
\caption{Comparison of Evaluation Metrics Performance}
\label{tab:metrics_sorted}

\resizebox{\columnwidth}{!}{
\begin{tabular}{
  l
  S[table-format=2.3]
  S[table-format=1.3]
  S[table-format=1.3]
  S[table-format=-1.3]
  S[table-format=1.3]
}
\toprule
\textbf{Metric} & 
\textbf{Effect} & 
\textbf{AUC} & 
\textbf{Genuine} & 
\textbf{Random} & 
\textbf{Mono} \\
\midrule
$AUC\text{-}PR$ & 38.990 & 0.964 & 0.714 & 0.057 & 0.885 \\
$AUC\text{-}ROC$ & 24.908 & 0.918 & 0.873 & 0.505 & 0.913 \\
$\mathit{Best\text{-}PwF}$ & 22.498 & 0.952 & 0.722 & 0.110 & 0.888 \\
$VUS^l\text{-}ROC$ & 19.383 & 0.907 & 0.865 & 0.510 & 0.908 \\
$PwF_{\beta}$ & 18.980 & 0.956 & 0.690 & 0.057 & 0.868 \\
$P@K$ & 18.916 & 0.951 & 0.688 & 0.057 & 0.860 \\
$TTF_\beta^\tau$ & 17.462 & 0.943 & 0.755 & 0.134 & 0.805 \\
$\mathit{K\%\text{-}PAF}^k_{\beta}$ & 17.050 & 0.967 & 0.741 & 0.077 & 0.865 \\
$VUS^l\text{-}PR$ & 14.707 & 0.938 & 0.668 & 0.061 & 0.908 \\
$CF_\beta$ & 10.425 & 0.926 & 0.694 & 0.092 & 0.868 \\
$PATE^{\varepsilon\delta}(F1)$ & 8.340 & 0.895 & 0.645 & 0.163 & 0.923 \\
${dT\text{-}PAF}^k_{\beta}$ & 7.302 & 0.882 & 0.728 & 0.155 & 0.825 \\
$eTaF_\beta$ & 7.162 & 0.913 & 0.624 & 0.044 & 0.860 \\
$\mathit{LSF}^w_\beta$ & 6.184 & 0.919 & 0.797 & 0.292 & 0.702 \\
$TaF_\beta^\delta$ & 5.766 & 0.939 & 0.593 & 0.057 & 0.815 \\
$RF^{\alpha}_{\beta}$ (front) & 5.653 & 0.938 & 0.605 & 0.073 & 0.867 \\
$RF^{\alpha}_{\beta}$ (flat) & 5.569 & 0.940 & 0.604 & 0.073 & 0.868 \\
$TD$ & 4.342 & 0.913 & 0.904 & 0.896 & 0.440 \\
$SF_{\beta}$ & 4.044 & 0.912 & 0.559 & 0.051 & 0.689 \\
$AF_\beta$ & 3.921 & 0.933 & 0.840 & 0.593 & 0.832 \\
$PAF_{\beta}$ & 3.715 & 0.930 & 0.842 & 0.433 & 0.653 \\
$NAB$ & -0.420 & 0.645 & 0.297 & 0.293 & -0.552 \\
\bottomrule
\end{tabular}
}

\endgroup
\end{table}

\subsection{Experimental Results}
To intuitively illustrate the distributional differences, Table~\ref{tab:metrics_sorted} and Figures~\ref{fig:sensitivity_heatmap}–\ref{fig:effect_auc} summarize the main quantitative and visual results, including: (1) score distributions ranked by composite performance, and (2) box–scatter plots jointly comparing effect size and AUC. The guiding principle is as follows: if a metric exhibits large, stable, and consistent differences between the score distributions of “genuine detectors” and “random predictors,” it can be considered a valid reflection of detection quality; otherwise, such metrics should be used cautiously or only as part of a composite evaluation framework.

All experiments were independently repeated five times to reduce the impact of sample randomness. In the boxplots, the green boxes (representing genuine detectors) are consistently and significantly higher than the red boxes (random predictions), with minimal overlap, indicating that random guessing almost never achieves comparable scores to genuine models. These results demonstrate that most metrics can reliably and significantly distinguish genuine detection from random guessing.

\begin{figure*}[!htbp]
    \centering
    \includegraphics[width=0.9\linewidth]{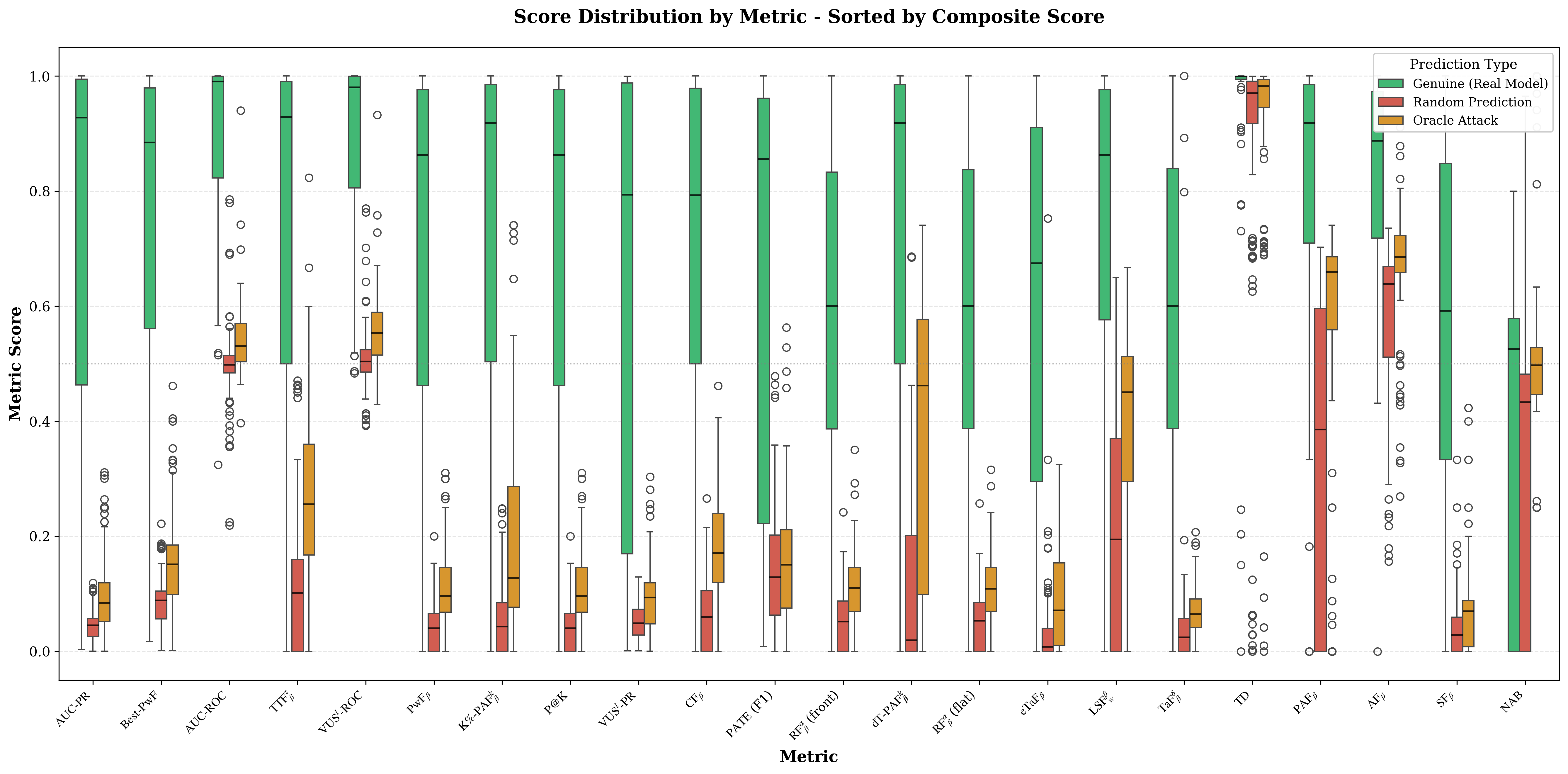}
        \begingroup

    \caption{Distribution of metric scores under genuine detectors, random guessing, and oracle-based attacks. 
    Metrics whose genuine and random score distributions largely overlap are considered less robust and should be used cautiously or in combination with other stable metrics. 
    Note that since the score ranges of TD and NAB differ from typical F1-based metrics (not bounded within [0, 1]), 
    the NAB scores are normalized to [0, 1], and the TD scores are inversely normalized because smaller distances indicate better performance.}
    \endgroup
    \label{fig:distribution}
\end{figure*}

        \begingroup

In contrast, several metrics exhibit noticeably weaker discriminative ability, including $PAF_\beta$, $AF_\beta$, $SF_\beta$, $TD$, and especially $NAB$. 
For example, although $PAF_\beta$ still produces relatively high scores for genuine detectors, the average scores of genuine and random predictors remain close (0.842 vs. 0.433), resulting in a modest effect size of only 3.715 and an AUC of 0.930. 
A similar pattern can be observed for $AF_\beta$, where the genuine and random scores (0.840 vs. 0.593) are insufficiently separated despite a high average genuine score.

The issue becomes more pronounced for $TD$, where the average scores of genuine and random predictors are almost indistinguishable (0.904 vs. 0.896). 
Although the AUC remains relatively high (0.913), the effect size drops to 4.342 and the monotonicity coefficient decreases substantially ($\rho = 0.440$), indicating limited ability to consistently rank genuine detectors above random predictions.

The most problematic case is $NAB$, which exhibits both a very low AUC (0.645) and a negative effect size (-0.420), along with a negative monotonicity coefficient ($\rho = -0.552$). 
This indicates that, under the random baseline setting, the metric may even rank random predictions comparably to or better than genuine detectors. 
Consequently, these metrics should be interpreted with caution. In practice, they are better used alongside more robust metrics or avoided as standalone evaluation criteria when assessing anomaly detection performance.

\endgroup

Additionally, as illustrated by the orange distributions in Figure~\ref{fig:distribution}, we also present the score distributions under \emph{oracle attack} conditions (serving as a stress test to reveal potential “gaming” upper bounds). Overall, several metrics—particularly those sensitive to anomaly length or boundary coverage—can approach genuine detector scores under oracle optimization. This suggests that, in practice, certain metrics may be artificially inflated through targeted optimization. Although the oracle results are not used as formal evaluation evidence, they serve as an essential indicator of potential robustness risks.

\begingroup

At the same time, we also took into account the impact of the dataset situation on this experiment, so we conducted a sensitivity analysis on the dataset situation as well. Several clear patterns emerge from Figure~\ref{fig:sensitivity_heatmap}. 
First, the discriminative ability of most metrics exhibits the strongest dependence on the signal-to-noise ratio (SNR). Both separation indicators show consistently positive correlations with SNR, often in the range of $\phi \approx 0.4$–$0.7$ with strong statistical significance. This indicates that metrics more clearly distinguish genuine detectors from random predictions when anomaly scores are well separated from normal scores, which corresponds to easier detection scenarios.

In contrast, the influence of anomaly density is generally weak. For most metrics, the correlation magnitude remains below $|\rho|<0.3$ and rarely reaches statistical significance. This suggests that the relative ranking of detectors produced by these metrics is largely stable across datasets with different contamination levels.

A similar pattern is observed for time-series length. The correlations between series length and metric separation are typically small and often statistically insignificant, indicating that the discriminative behavior of most metrics does not strongly depend on the temporal scale of the dataset.

Importantly, these trends are consistent across both separation indicators (Effect Size and AUC), providing additional evidence that the observed patterns are robust. Overall, the results suggest that while detection difficulty (captured by SNR) naturally influences the strength of metric separation, the proposed evaluation analysis remains largely stable with respect to dataset scale and anomaly density across the benchmark datasets.

Moreover, it is important to note that the benchmark datasets used in this study cover a broad spectrum of detection difficulty. 
In addition to synthetic datasets with relatively clear anomaly patterns, the benchmark also includes real-world datasets that often exhibit substantially lower signal-to-noise ratios. 
This diversity ensures that the analysis captures both low-noise and high-noise regimes. 
Consequently, the observed dependence on SNR primarily reflects intrinsic task difficulty rather than bias toward a particular dataset type, further supporting the robustness of the experimental conclusions.

\textbf{Sensitivity to model type.} Beyond dataset characteristics, we examine how the choice of anomaly detection model affects the separation results. Figure~\ref{fig:sensitivity_heatmap} presents the mean $z$-score for each metric--model combination, aggregated over all benchmark datasets.

Several observations stand out. First, all four model types achieve positive $z$-scores for the vast majority of metrics, with overall means of $10.8$ (AE), $9.7$ (AR), $9.1$ (TimesNet), and $6.6$ (LOF). This confirms that the evaluation framework reliably distinguishes genuine detectors from random predictions regardless of the underlying model paradigm.

Second, while the absolute $z$-scores differ across models, the \emph{ordinal ranking} of metrics is remarkably consistent. AE achieves the highest mean separation for 21 out of 23 metrics, and LOF ranks last for 20 out of 23 metrics, yet the relative ordering among the remaining metrics is largely preserved. This indicates that metrics identified as having strong (or weak) discriminative ability retain that characterization across different detector families.

Third, we test whether metric separation increases monotonically with model complexity by computing the Spearman correlation between an ordinal complexity encoding (AR$<$LOF$<$TimesNet$<$AE) and the per-metric $z$-score. None of the 23 metrics yields a statistically significant correlation ($p<0.05$), with $\rho$ values ranging from $-0.14$ to $+0.17$. This confirms that the evaluation conclusions do not systematically favor simpler or more complex models.

Finally, only two metrics---NAB and TD---exhibit negative mean $z$-scores across all model types, suggesting that random predictions occasionally outscore genuine detectors under these criteria. This is consistent with their known sensitivity to specific temporal scoring assumptions and provides further evidence for treating them with caution in practical benchmarking.

Overall, these results demonstrate that the metric evaluation conclusions drawn in this study are robust not only to dataset-level characteristics (anomaly density, series length, SNR) but also to the choice of anomaly detection model, reinforcing the generalizability of the proposed evaluation framework.

\endgroup

\begin{figure*}[!h]
    \centering
    \includegraphics[width=0.9\linewidth]{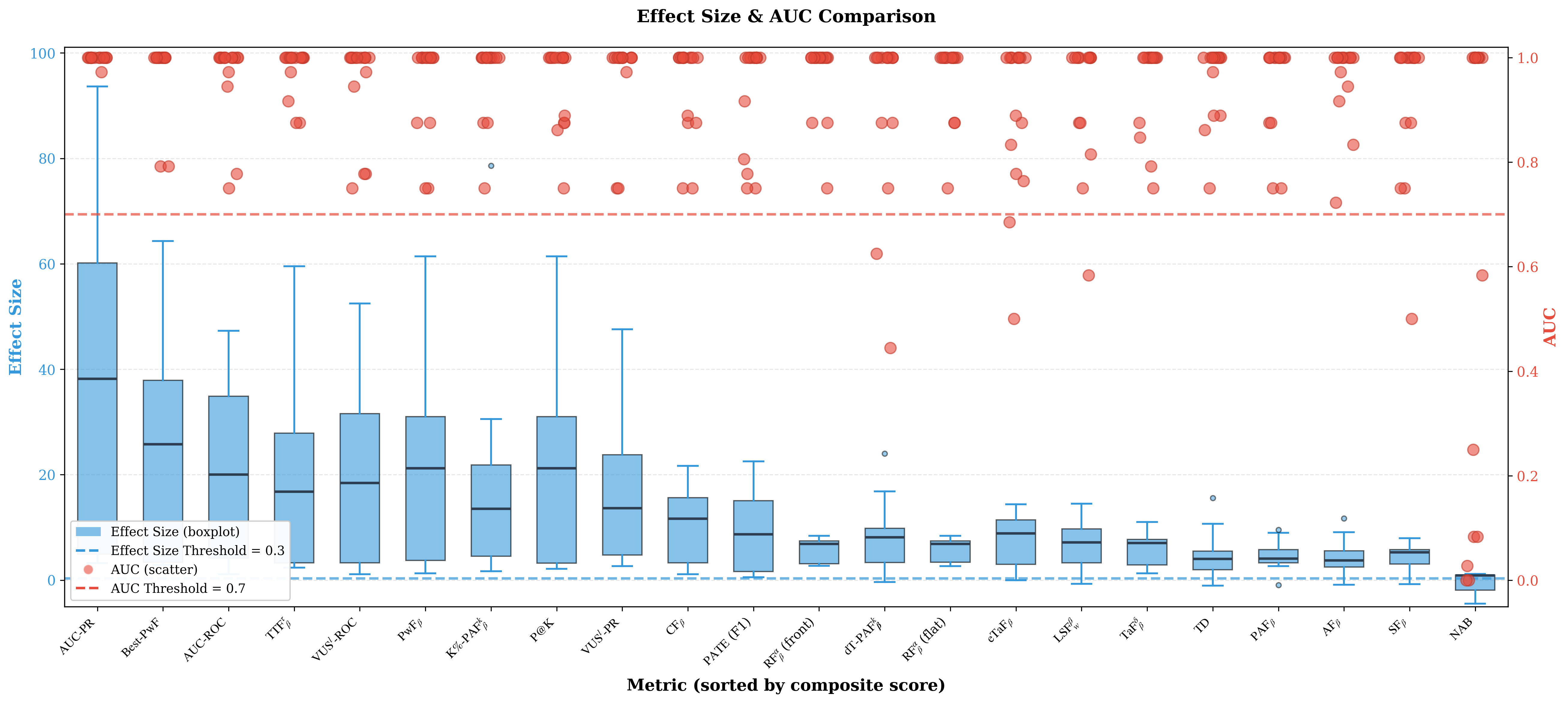}
    \begingroup

    \caption{Joint analysis of effect size and AUC across all metrics. Metrics located in the left region exhibit high discriminative ability, clearly separating genuine and random scores, while those near the right indicate weak robustness to random guessing.}
    \endgroup
   \label{fig:effect_auc}
\end{figure*}

\begin{figure*}[!h]
    \centering
    \includegraphics[width=\linewidth]{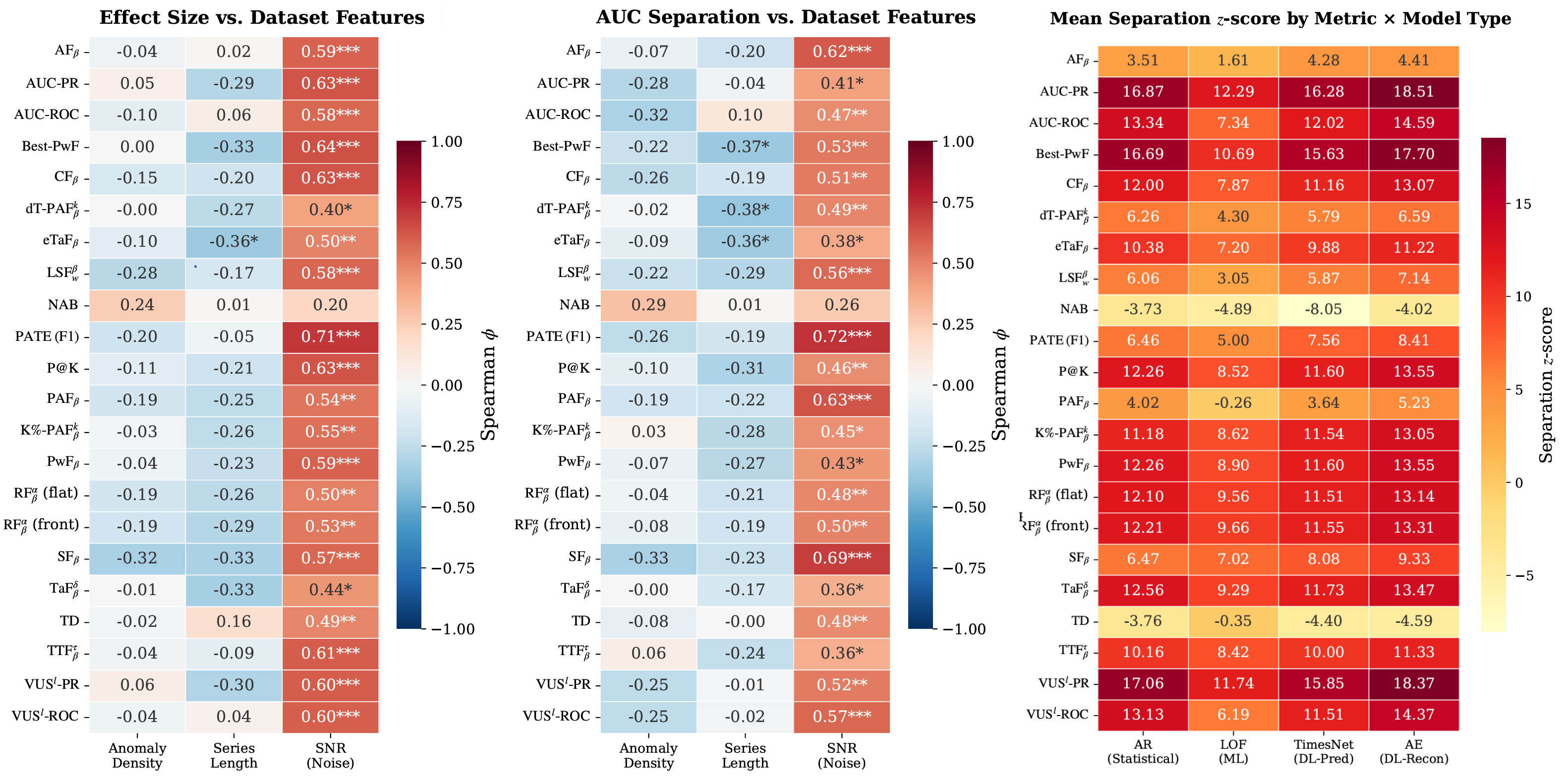}

    \begingroup

    \caption{
    Sensitivity analysis of metric discriminative ability with respect to dataset characteristics. 
    Each cell reports the Spearman rank correlation coefficient ($\phi$) between a dataset feature (anomaly density, time-series length, or signal-to-noise ratio) and the metric's separation performance across all dataset–curve instances. 
    Two separation indicators are shown: Effect Size (left) and AUC-based separation (right). 
    Values close to zero indicate that the metric's discriminative ability is largely insensitive to the corresponding dataset property, whereas larger magnitudes suggest stronger dependence. 
    Statistical significance is indicated by $^{*}p<0.05$, $^{**}p<0.01$, and $^{***}p<0.001$.
    }

    \endgroup
    \label{fig:sensitivity_heatmap}
\end{figure*}

\begin{table*}[t]
\centering
\small
\begingroup

\caption{Capability Matrix of Representative TSAD Evaluation Metrics}

\label{tab:metric_capability}

\begin{tabular}{lccccc}
\toprule
Metric & Timeliness Reward & Label Tolerance & Cost Awareness & Random Robustness & Parameter-Free \\
\midrule
$\text{dT-PAF}^{k}_\beta$ & \checkmark &        &            &  \checkmark  &            \\
$\text{RF}^{\alpha}_\beta$ & \checkmark &       &            &  \checkmark  &            \\
$\text{NAB}$          & \checkmark &            & \checkmark &            &            \\
$\text{LSF}^{w}$      & \checkmark &            & \checkmark & \checkmark &            \\
\midrule
$\text{TTF}^{\tau}_\beta$   &       & \checkmark &            & \checkmark &            \\
$\text{TaF}^{\delta}_\beta$ &      & \checkmark &            & \checkmark &            \\
$\text{eTaF}_\beta$   &            & \checkmark &            & \checkmark &            \\
$\text{AF}_\beta$     &            & \checkmark &            &         & \checkmark \\
$\text{TD}$           &            & \checkmark &            &        & \checkmark \\
$\text{VUS}^{l}$      &            & \checkmark &            &  \checkmark  &  \\
$\text{PATE}^{\varepsilon\delta}$ & \checkmark & \checkmark & & \checkmark  &            \\
\midrule
$\text{PAF}_\beta$ &       &            &            &  &            \\

$\text{P@K}$          &            &            & \checkmark &  \checkmark &  \\

$\text{K\%-PAF}^{k}_\beta$ &       &            &            & \checkmark &            \\

$\text{Best-PwF}$     &            &            &            &  \checkmark  & \checkmark \\
$\text{AUC}$          &            &            &            & \checkmark & \checkmark \\
\bottomrule

\end{tabular}
\endgroup
\end{table*}

\begin{table*}[t]
\centering
\small
\begingroup

\caption{Recommended Default Metric Sets for Common TSAD Evaluation Scenarios }

\label{Recommended Default Metric Sets}
\begin{tabular}{p{2.35cm} p{2.1cm} p{12.4cm}}
\toprule
\textbf{Scenario} & \textbf{Recommended Metrics} & \textbf{Rationale} \\
\midrule

General benchmarking &
$\text{K\%-PAF}^{k}_\beta$+ $\text{VUS}^{l}$/$\text{AUC}$ &
$\text{K\%-PAF}^{k}_\beta$ evaluates segment-level overlap and detection completeness, capturing structural correctness of predicted anomaly segments. 
$\text{VUS}^{l}$/$\text{AUC}$ complements this by integrating performance across thresholds, providing a ranking-oriented perspective that improves robustness when models produce different score distributions. \\

Timeliness-critical detection &
PATE$^{\varepsilon\delta}$+ RF &
PATE models proximity to anomaly boundaries with graded rewards, explicitly capturing early-warning capability. 
Range-based F-score preserves structural overlap evaluation of detected segments. 
Compared with reward-curve-based alternatives such as NAB, this combination offers clearer interpretability and less sensitivity to reward-function design. \\

Label-uncertain datasets &
$\text{VUS}^{l}$ + $\text{eTaF}_\beta$ &
$\text{VUS}^{l}$ introduces tolerance-based robustness to boundary misalignment by aggregating performance across multiple tolerance levels. 
$\text{eTaF}_\beta$ further decomposes detection quality into event-level recognition and positional accuracy, improving interpretability when annotation boundaries are uncertain. 
Distance-only formulations (e.g., TD or AF) may emphasize geometric proximity while being less stable under random scoring patterns. \\

Human verification cost sensitive &
$\text{LSF}^{w}$ + P@K &
$\text{LSF}^{w}$ models the temporal clustering of false alarms and converts dispersed false positives into verification workload units. 
P@K captures analyst capacity constraints by evaluating whether the most critical alerts appear among the highest-ranked predictions. 
Together they represent two complementary operational perspectives: verification workload and review budget prioritization. \\

Threshold-free model comparison &
$\text{AUC}$ + Best-PwF &
AUC evaluates ranking robustness by integrating performance across the entire threshold space, enabling fair comparison of detectors with different score calibrations. 
Best-PwF provides an upper-bound reference corresponding to optimal threshold selection, helping interpret the achievable performance gap between ranking quality and deployable operating points. \\

\bottomrule

\end{tabular}
\endgroup
\end{table*}

\begingroup

\section{Practical Guidance for Metric Selection}

The previous sections systematically analyzed the different evaluation requirements addressed by time series anomaly detection (TSAD) metrics from a problem-oriented perspective. However, in practical research and industrial applications, practitioners often face a more direct question:
\textit{Which evaluation metric should be chosen for a specific task scenario?}

Since different metrics focus on different evaluation objectives, their suitability varies significantly across application contexts. For example:

\begin{itemize}
\item In \textbf{fault prediction systems}, detecting anomalies as early as possible is critical; therefore, metrics that provide \textbf{timeliness rewards} are desirable.

\item In \textbf{industrial monitoring systems}, the start and end boundaries of anomaly labels are often uncertain; thus, metrics that are tolerant to \textbf{label imprecision} are preferred.

\item In scenarios with \textbf{high manual inspection costs} (e.g., medical diagnosis or industrial maintenance), metrics should reflect the cost associated with \textbf{false alarms}.

\item In \textbf{benchmark comparisons across algorithms}, metrics that are \textbf{parameter-free} and highly comparable are generally preferred.
\end{itemize}

Therefore, metric selection can be viewed as a problem of \textbf{matching evaluation objectives with the capabilities of evaluation metrics}.

To help researchers quickly understand the capability characteristics of existing metrics, Table~\ref{tab:metric_capability} summarizes the capabilities of representative TSAD evaluation metrics across several evaluation dimensions.

\subsection{Capability Matrix of Existing Metrics}

Table~\ref{tab:metric_capability} presents a capability matrix of representative TSAD metrics, where each column corresponds to a specific evaluation capability.

It is important to note that most metrics are not designed to address only a single issue. Instead, many of them possess multiple capabilities simultaneously. However, these capabilities are often implemented through additional parameters or heuristic rules. As a result, there is often a trade-off between \textbf{benchmark comparability} and \textbf{task-specific adaptability} when selecting evaluation metrics.

\subsection{How to Choose Metrics for Different Tasks}

Based on the capability analysis above, we suggest the following practical guidelines when selecting evaluation metrics.

\textbf{(1) Algorithm benchmark evaluation}

If the goal is to pursue \textbf{strictly fair performance comparison across models}, it is preferable to adopt \textbf{parameter-free metrics}, such as AUC and $\text{Best-PwF}$, which avoid evaluation bias introduced by thresholds or window parameters.

Otherwise, a practical default combination is $\text{K\%-PAF}^{k}_\beta$ + $\text{VUS}^{l}$/$\text{AUC}$. 
$\text{K\%-PAF}^{k}_\beta$ captures segment-level detection quality, while $\text{VUS}^{l}$/$\text{AUC}$ complements it with a threshold-robust ranking perspective.

\textbf{(2) Early warning systems}

For tasks emphasizing \textbf{early detection of anomalies}, metrics incorporating \textbf{timeliness rewards} should be adopted, such as:
$\text{dT-PAF}^{k}_\beta$, $\text{RF}^{\alpha}_\beta$, $\text{LSF}^{w}$.

These metrics assign higher scores to earlier detections and therefore better reflect the early-warning capability of anomaly detection systems.

\textbf{(3) Scenarios with imprecise labels}

In many industrial and real-world datasets, the exact boundaries of anomalies are often uncertain. In such cases, it is preferable to adopt metrics that are tolerant to label imprecision, including: $\text{TTF}^{\tau}_\beta$, $\text{TaF}^{\delta}_\beta$ , $\text{eTaF}_\beta$, $\text{VUS}^{l}$, $\text{PATE}^{\varepsilon\delta}$.

These metrics reduce evaluation bias caused by annotation inaccuracies.

\textbf{(4) High manual inspection cost scenarios}

In applications such as healthcare, industrial maintenance, or spacecraft monitoring, each alarm may trigger manual inspection. Since these systems often require domain experts to verify alarms rather than general operators, the cost of manual inspection can be substantial. Therefore, metrics that incorporate \textbf{false alarm penalties} or \textbf{sparse alarm constraints} are more suitable, such as: $\text{LSF}^{w}$, $\text{P@K}$.

These metrics better reflect the operational cost structure of real-world monitoring systems.

While many evaluation metrics have been proposed for TSAD, no single metric can fully capture all relevant aspects of anomaly detection performance. 
Therefore,as shown in Table~\ref{Recommended Default Metric Sets}, instead of recommending a single universal metric, we suggest using small combinations of complementary metrics as default evaluation sets for common scenarios.

The recommended metric sets follow three guiding principles.

First, each combination integrates different evaluation perspectives. 
For example, range-based metrics focus on structural overlap between predicted and ground-truth anomaly segments, tolerance-based metrics capture robustness to boundary uncertainty, and ranking-based metrics evaluate the quality of anomaly score ordering. 
Using multiple metrics therefore reduces evaluation bias introduced by any single modeling assumption.

Second, the recommended sets prioritize metrics that exhibit relatively stable behavior under noisy or weak predictors. 
Certain formulations that rely heavily on geometric proximity or specific reward-curve designs may produce unstable scores when models generate random or highly noisy anomaly scores. 
To maintain practical reliability, the proposed combinations emphasize metrics that balance interpretability, robustness, and structural relevance.

Third, the selection aims to balance theoretical expressiveness with practical usability. 
Some metrics introduce extensive parameterization or computational overhead, which may limit their applicability in routine benchmarking. 
The recommended combinations therefore focus on metrics that provide complementary information while remaining interpretable and computationally tractable.

These recommended sets should be interpreted as practical starting points rather than strict standards. 
Researchers and practitioners may adapt the metric choices according to domain-specific requirements, annotation characteristics, and operational constraints.

\subsection{Implications for Future Metric Design}

The above analysis also provides several important implications for the design of future evaluation metrics.

First, different application scenarios impose different evaluation requirements. As a result, \textbf{no single metric is capable of satisfying all evaluation objectives simultaneously}.

Future metric design may consider the following directions.

\textbf{(1) Modular capability design}

An ideal evaluation metric should allow different capability modules to be combined, such as:

\begin{itemize}
\item timeliness reward
\item label tolerance
\item cost-aware penalty
\end{itemize}

Such modular design would allow metrics to be flexibly configured according to the requirements of different tasks.

\textbf{(2) Balancing parameterization and comparability}

Many capabilities (e.g., timeliness reward or label tolerance) inevitably require the introduction of parameters, such as: time decay functions, tolerance window sizes, cost weights. Therefore, completely \textbf{parameter-free metrics} are often insufficient to capture complex evaluation requirements.

In practice, future evaluation frameworks may adopt a \textbf{dual-track structure}:

\begin{itemize}
\item \textbf{Parameter-free metrics}, which are suitable for fair benchmark comparisons.

\item \textbf{Capability-aware metrics}, which are designed for task-specific evaluation in real-world applications.
\end{itemize}

\textbf{(3) Improving robustness against random predictions}

Experimental results show that some widely used metrics (e.g., Point-Adjust or NAB) may still obtain relatively high scores under random predictions. Future metric design should therefore pay greater attention to:

\begin{itemize}
\item penalizing random predictions
\item suppressing opportunistic or sparse prediction strategies
\end{itemize}

\textbf{(4) Periodic re-evaluation of framework conclusions} 

As the TSAD field evolves, metric discriminability and evaluation priorities may shift with new models and datasets. We therefore recommend periodically re-running the established experimental pipeline—covering genuine detector comparisons, random baselines, and oracle stress tests—to verify the stability of metric rankings. This requires no framework redesign; the protocol itself serves as a living evaluation tool, enabling the community to detect shifts in metric suitability early and update selection guidelines accordingly.

This will help prevent misleading evaluation results and improve the reliability of performance comparisons.

\section{Conclusion}

This study presents a problem-oriented framework for evaluating 
time series anomaly detection metrics, classifying over twenty 
widely used metrics into six functional dimensions---basic 
evaluation, timeliness rewards, tolerance to annotation 
uncertainty, human-audit cost penalties, robustness against 
random scoring, and parameter-free comparability---according 
to the underlying evaluation problems they address rather than 
their mathematical form.

Through this perspective, we uncover that many evaluation 
inconsistencies originate from mismatched design intentions 
rather than mathematical deficiencies, and that metric 
suitability is inherently task-dependent. Systematic robustness 
experiments further demonstrate that metrics differ by up to an 
order of magnitude in their ability to distinguish genuine 
detectors from random guessing, and that widely adopted metrics 
such as NAB and $\text{PAF}_\beta$ show critically limited 
resistance to random-score inflation.

By rethinking metric evaluation from a problem-driven standpoint, 
this work contributes a unified perspective that clarifies 
existing metric diversity and inspires the principled design of 
more context-aware, equitable, and operationally meaningful 
evaluation systems for time series anomaly detection.

\printcredits

\bibliographystyle{elsarticle-num}

\bibliography{references}



\end{document}